\documentclass[11pt,review,a4paper,fleqn]{cas-sc}

\usepackage[authoryear]{natbib}
\usepackage{hyperref}
\hypersetup{colorlinks=true, citecolor=blue, linkcolor=blue, urlcolor=blue}

\usepackage{tabularx}
\usepackage{lineno}
\usepackage{booktabs}
\usepackage{multirow}
\usepackage{graphicx}
\usepackage{booktabs}       
\usepackage{threeparttable} 
\usepackage{ulem}
\usepackage{caption}
\usepackage{cleveref}
\usepackage{float}
\usepackage{subcaption}  
\usepackage{amsmath}

\def\tsc#1{\csdef{#1}{\textsc{\lowercase{#1}}\xspace}}
\tsc{WGM}
\tsc{QE}
\tsc{EP}
\tsc{PMS}
\tsc{BEC}
\tsc{DE}

\begin{document}
\let\printorcid\relax
\let\WriteBookmarks\relax
\def\floatpagepagefraction{1}
\def\textpagefraction{.001}
\shortauthors{Z. Song et~al.}
This work has been submitted to the KeAi for possible publication. Copyright may be transferred without notice, after which this version may no longer be accessible
\title [mode = title]{TCLeaf-Net: a transformer-convolution framework with global-local attention for robust in-field lesion-level plant leaf disease detection}                      

\author[1]{Zishen~Song}[type=editor,auid=000,bioid=1]
\fnmark[1]
\credit{Conceptualization, Methodology, Software, Formal analysis, Writing - original draft, Writing - review \& editing}

\author[2]{Yongjian~Zhu}
\fnmark[1]
\credit{Conceptualization, Methodology, Formal analysis, Writing - original draft, Funding acquisition, Writing - review \& editing}

\author[3]{Dong~Wang}
\cormark[1]
\credit{Project administration, Resources, Data curation, Writing - review \& editing, Supervision}

\author[1]{Hongzhan~Liu}
\cormark[1]
\credit{Resources, Funding acquisition, Supervision, Writing - review \& editing}

\author[4]{Lingyu~Jiang}
\credit{Validation, Formal analysis, Writing - review \& editing}

\author[3]{Yongxing~Duan}
\credit{Data curation, Writing - review \& editing}

\author[3]{Zehua~Zhang}
\credit{Investigation, Writing - review \& editing}

\author[3]{Sihan~Li}
\credit{Visualization, Writing - review \& editing}

\author[3]{Jiarui~Li}
\credit{Validation, Writing - review \& editing}

\fntext[fn1]{These authors contributed equally to this work.}
\cortext[cor1]{Corresponding authors: dongwang@sit.edu.cn (D.~Wang); lhzscnu@163.com (H.~Liu).}

\affiliation[1]{organization={School of Optoelectronic Science and Engineering, South China Normal University},
                city={Guangzhou},
                postcode={510006},
                state={Guangdong},
                country={China}}

\affiliation[2]{organization={College of Engineering Physics, Shenzhen Technology University},
                city={Shenzhen},
                postcode={518118},
                state={Guangdong},
                country={China}}

\affiliation[3]{organization={Faculty of Intelligent Technology, Shanghai Institute of Technology},
                city={Shanghai},
                postcode={201418},
                state={Shanghai},
                country={China}}

\affiliation[4]{organization={Graduate School of Information Sciences, Tohoku University},
                city={Sendai},
                postcode={9808579},
                state={Miyagi},
                country={Japan}}

\begin{abstract}
Timely and accurate detection of foliar diseases is vital for safeguarding crop growth and reducing yield losses. Yet, in real-field conditions, cluttered backgrounds, domain shifts, and limited lesion-level datasets hinder robust modeling. To address these challenges, we release Daylily-Leaf, a paired lesion-level dataset comprising 1,746 RGB images and 7,839 lesions captured under both ideal and in-field conditions, and propose TCLeaf-Net, a transformer-convolution hybrid detector optimized for real-field use. TCLeaf-Net is designed to tackle three major challenges. To mitigate interference from complex backgrounds, the transformer-convolution module (TCM) couples global context with locality-preserving convolution to suppress non-leaf regions. To reduce information loss during downsampling, the raw-scale feature recalling and sampling (RSFRS) block combines bilinear resampling and convolution to preserve fine spatial detail. To handle variations in lesion scale and feature shifts, the deformable alignment block with FPN (DFPN) employs offset-based alignment and multi-receptive-field perception to strengthen multi-scale fusion. Experimental results show that on the in-field split of the Daylily-Leaf dataset, TCLeaf-Net improves mAP@50 by 5.4 percentage points over the baseline model, reaching 78.2\%, while reducing computation by 7.5 GFLOPs and GPU memory usage by 8.7\%. Moreover, the model outperforms recent YOLO and RT-DETR series in both precision and recall, and demonstrates strong performance on the PlantDoc, Tomato-Leaf, and Rice-Leaf datasets, validating its robustness and generalizability to other plant disease detection scenarios.
\end{abstract}





\maketitle

\section{Introduction}

Agriculture is a cornerstone of the global economy, yet climate change and ecological degradation are elevating the incidence of foliar diseases, with consequent yield losses and economic damage \citep{ristaino2021persistent}. Leaf-borne pathogens reduce the photosynthetically active area, disrupt transpiration, and trigger systemic stress responses, so even small, early-stage lesions can translate into substantial yield penalties if not treated promptly. Early warning therefore hinges on lesion-level monitoring, where individual spots are detected before they coalesce into whole-leaf infections.

Daylily cultivation, for example, has expanded rapidly in Yunzhou, China, generating 75,000-105,000 CNY per hectare and contributing significantly to poverty alleviation \citep{land13040439}. However, abnormal temperature and humidity have increased the probability of disease outbreaks, eroding these gains \citep{wei2022first}. Reliable, in-field, lesion-level detection is thus essential for timely chemical or biological intervention and for sustaining the economic benefits of high-value crops such as daylily.

These challenges highlight the need for disease detection and intervention. Environmental sensors (e.g., temperature, humidity) monitor plant health at the macro level but cannot localize lesions and are sensitive to deployment conditions. In contrast, low-cost RGB cameras on GPS-enabled platforms provide flexible on-site lesion localization. This generates large volumes of images requiring an automated analysis method.

Traditional methods such as SVM, ANN, and decision trees rely on hand-crafted features with classifiers. However, these approaches are insufficiently robust under cluttered field conditions and domain shifts \citep{khan2024automated,Banerjee2023RidgeGourd,Hossain2018TeaLeaf,joshi2022plant}. In recent years, CNN-based models have replaced manual feature engineering and improved robustness to background interference \citep{Farjon2023CountingReview}. They can learn hierarchical features directly from raw RGB images. Representative works include DGP-SNNet for fine-grained tomato disease discrimination \citep{Tiancan2024DGPSNNet}, mixed-data augmentation for robust rice leaf classification \citep{haikal2024comprehensive}, an improved ResNet50 for mildew detection in grapevine \citep{sagar2025precision}, the 28-layer DeepPlantNet \citep{Ullah2023DeepPlantNet}, and a ResNet50-MobileNetV2 ensemble for efficient tomato detection \citep{Bharathi2025TomatoEnsemble}  . These methods improved disease recognition accuracy through architectural improvements and data augmentation. However, the datasets used in these studies typically contain images in which the diseased leaf occupies a large portion of the image area and the background is either simple or only slightly complex. We refer to such datasets as ideal datasets. When these methods are applied to images captured under real-world conditions, where multiple leaves appear in a single frame and the background contains objects that visually resemble disease symptoms, which we refer to as in-field dataset, they often exhibit misclassification issues.

This limitation arises because convolutional models face constraints in receptive-field size and long-range context modeling \citep{zhang2020bridging}, which causes them to focus only on local regions such as specific lesions on leaves or background elements that visually resemble disease symptoms, while ignoring the semantic relationships between the disease areas and the leaf, or between the leaf and the background. Although increasing the network depth can theoretically enlarge the receptive field, it often comes at the cost of reduced feature-map resolution, which may result in the loss of important object details. To further address this issue and enhance the receptive field, transformer-based architectures have been introduced to capture global semantic dependencies and improve robustness across multiple scales.

Transformer-based models have recently attracted attention in plant disease detection. For example, CAST-Net~\citep{Zhang2024CASTNet} combined convolution and self-attention for leaf disease recognition, MobileH-Transformer~\citep{Thai2024MobileHTransformer} integrated convolutional downsampling with transformer layers to improve classification, and other models applied multi-head attention and depthwise separable convolution for potato disease classification~\citep{Zhang2024MDSCIRNet}. But these studies rarely explored why transformers enhance performance and why they have been evaluated mainly on ideal datasets.

The existing methods are trained on datasets such as PlantVillage, Turkey-PlantDataset, PlantDoc, and AES-CD9214. However, most of these datasets do not represent field complexity such as lighting variations, scale variations, occlusions, and background clutter\citep{attri2023review, ghazal2024computer}. Field images often include noisy backgrounds and unstable illumination, which can degrade model performance~\citep{liu2025deep}. Therefore, most research lacks sufficient validation under in-field conditions, limiting their generalization.

Most studies focus on whole-leaf classification or diseased leaf detection (leaf-level detection), but overlook lesion-level localization and classification. This results in leaves being categorized as a single type, reducing classification precision. While a few studies collected field data, most did not release their datasets, further limiting reproducibility.

To sum up, we identify four major limitations:
\begin{enumerate}
\item Most studies use clean, ideal-conditions datasets; no high-quality public object detection dataset reflects in-field foliar disease conditions.

\item Existing works focus on leaf-level classification or detection, neglecting lesion-level objects with fine-grained annotation.

\item CNNs have a limited effective receptive field and weak long-range dependency modeling; simply deepening the network enlarges the theoretical field but reduces feature-map resolution and erodes fine textures. By contrast, transformers incur quadratic attention complexity and exhibit weak spatial inductive bias, which can lead to diffuse attention under clutter and higher data requirements.

\item In-field scenarios, lesions vary in size, leading to intra-class feature inconsistency and complicating multi-scale fusion.
\end{enumerate}

Limitations 1 and 2 have also been highlighted in recent reviews~\citep{bhargava2024plant,jafar2024revolutionizing,simhadri2024deep}. Once addressed, the next challenge is tackling limitations 3 and 4.

To address these challenges, it is necessary to construct a high-quality in-field dataset with lesion-level annotations. Models should be able to suppress background noise effectively. The limitations of convolutional models underscore the need for transformer-based architectures. Prior works \citep{thai2023formerleaf,reis2024potato,chakrabarty2024interpretable,huang2025econv} have demonstrated that integrating convolution module with transformers can enhance plant disease classification. However, most of these studies employ single branch transformer architectures, which lead to large training data requirements. In addition, these models are primarily designed for ideal datasets; they either lack validation on in-field datasets or exhibit poor robustness when evaluated in real-world field conditions.

Multi-scale detection remains a significant challenge. Beyond the complications introduced by complex backgrounds, a major obstacle in in-field datasets is the variability in lesion scale. Techniques such as lateral feature fusion in feature pyramid networks (FPN) can enhance semantic richness, but they often suffer from feature misalignment, which impairs the detection of small lesions \citep{huang2021fapn,van2023feature}. The deformable convolutions employed in the feature alignment pyramid network (FaPN) \citep{liu2023tripartite} offer valuable inspiration, and we accordingly incorporate this strategy into our own framework.

To fill these research gaps, this study proposes TCLeaf-Net and publishes two datasets. TCLeaf-Net is a transformer-convolution hybrid model for detecting daylily leaf diseases in complex field conditions. It integrates a transformer-convolution module (TCM) that combines efficient attention (EA) for global semantics with convolutional layers for local texture and color, while reducing computational complexity.

Inspired by FaPN, we adopt its feature selector and, to provide richer contextual cues for the alignment block, design a multi-receptive-field perception (MRFP) module. These components constitute a deformable alignment block with FPN (DFPN) that mitigates feature misalignment in multi-scale fusion, while the detection head subsequently localizes lesion regions. The rationale for each component is detailed in Section~\ref{sec:TCLeaf}.

In addition, two lesion-level datasets, Daylily-Leaf (ideal) and Daylily-Leaf (in-field), are constructed and released, both with fine-grained bounding box annotations. Although modest in scale and coarsely categorized, they enable accurate lesion localization and provide a practical basis for early-stage classification. The datasets are also transferable to other crops and disease types, offering a valuable foundation for future fine-grained recognition and cross-task adaptation.

The main contributions of this paper are as follows:
\begin{enumerate}
\item This study builds and publishes two high-quality datasets, Daylily-Leaf (in-field) and Daylily-Leaf (ideal), supporting object detection research under real-world (in-field) and laboratory (ideal) conditions.

\item To focus on lesion areas and suppress background noise, this study combines convolution and transformer mechanisms and proposes a detection framework. The results demonstrate that this mechanism is effective in inhibiting background noise.

\item This study proposes a deformable alignment block with FPN (DFPN) to filter and align multi-scale features for better fusion, and introduce a raw-scale feature recalling and sampling (RSFRS) module to boost early semantic recall.

\item We empirically demonstrate that TCM mitigates transformer diffuse attention in cluttered field backgrounds, as evidenced by ablations and Grad-CAM analyses.

\end{enumerate}
\section{Method}

\subsection{The architecture of TCLeaf-Net}
\label{sec:TCLeaf}

\begin{figure}
 \centering
  \includegraphics[width=0.85\linewidth]{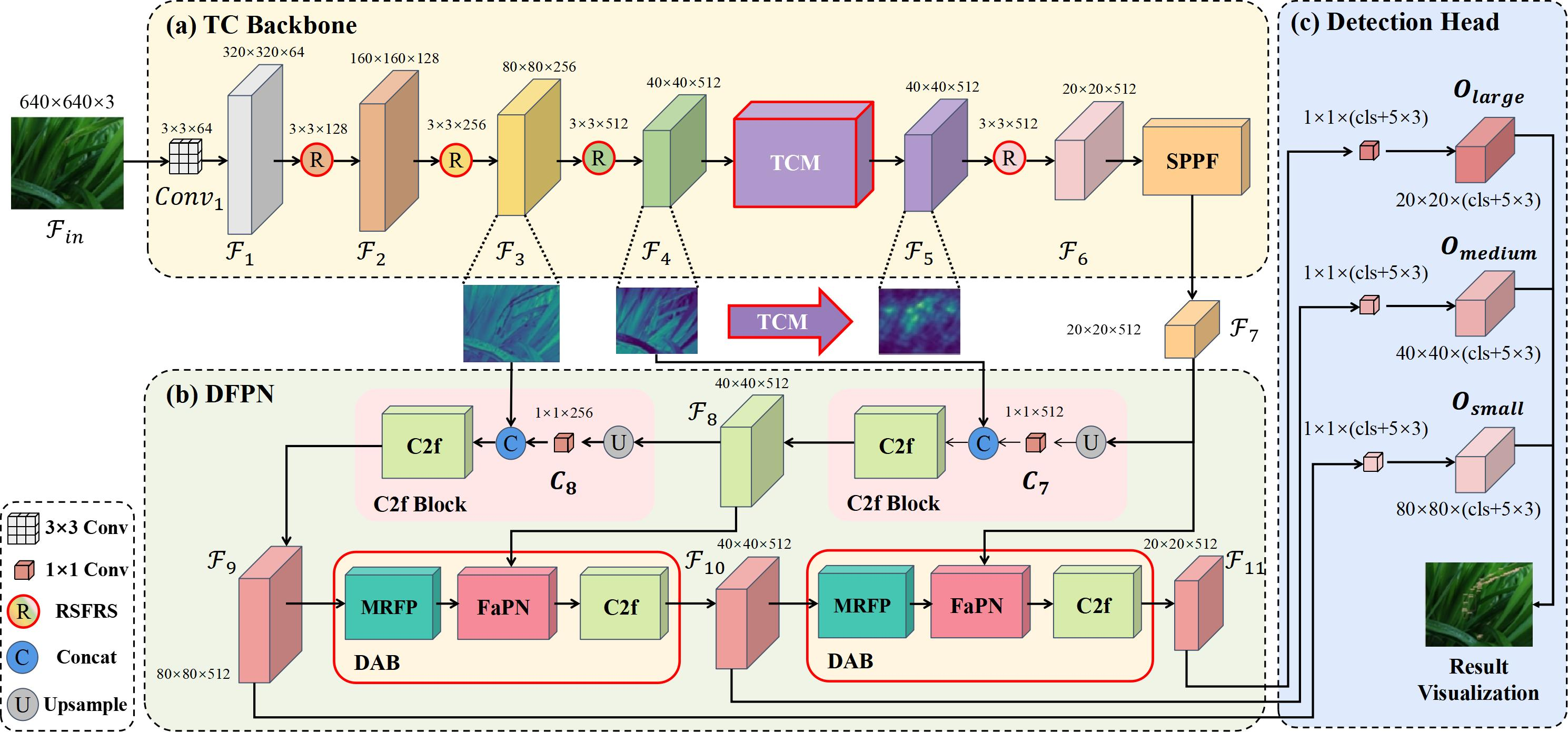}
  \caption{Overall architecture of TCLeaf-Net.  
  Modules proposed in this work are highlighted with bold red outlines.}
  \label{fig:Fig1}
\end{figure}

TCLeaf-Net mainly consists of a transformer-convolution hybrid backbone (TC backbone) and a deformable feature-alignment pyramid network (DFPN), followed by a decoupled detection head. As shown in Fig.~\ref{fig:Fig1}, the RGB image $F_{\mathrm{in}}$ is first fed into the TC backbone to produce multi-scale features $F_{3}, F_{4}, F_{7}$, where TCM suppresses non-leaf background and mitigates the diffuse-attention tendency of pure transformers, and RSFRS adaptively combines bilinear and convolutional downsampling to retain pre-downsampling spatial detail. The features are then input to DFPN for offset-based alignment and multi-receptive-field fusion, yielding $F_{9}, F_{10}, F_{11}$. Finally, three scale-specific decoupled heads perform classification and bounding-box regression, and the final detections are obtained after confidence thresholding and non-maximum suppression (NMS). The following sections describe each part in detail.

\subsection{TC backbone}
Given an RGB image $F_{\mathrm{in}}$ $\in\mathbb{R}^{640\times640\times3}$, the backbone applies a small-step overlapping patch embedding (SSOPE) layer, a $3\times3$ convolution with stride 2, to downsample the input while maintaining local continuity, producing $F_{1}\in\mathbb{R}^{320\times320\times64}$. A raw-scale feature recalling and sampling (RSFRS) module then carries out a second downsampling stage and intra-scale interaction via interpolation, yielding three resolutions: $F_{2}$ at $160\times160$, $F_{3}$ at $80\times80$, and $F_{4}$ at $40\times40$.
To encode both local context and long-range relations, $F_{4}$ is passed to a TCM, which outputs $F_{5}\in\mathbb{R}^{40\times40\times512}$. RSFRS is then applied to downsample $F_{5}$ to $F_{6}\in\mathbb{R}^{20\times20\times512}$, and a combination of SPPF further transforms $F_{6}$ into $F_{7}\in\mathbb{R}^{20\times20\times512}$, which is particularly informative for large-object detection.



\begin{figure}
  \centering
  \includegraphics[width=0.85\linewidth]{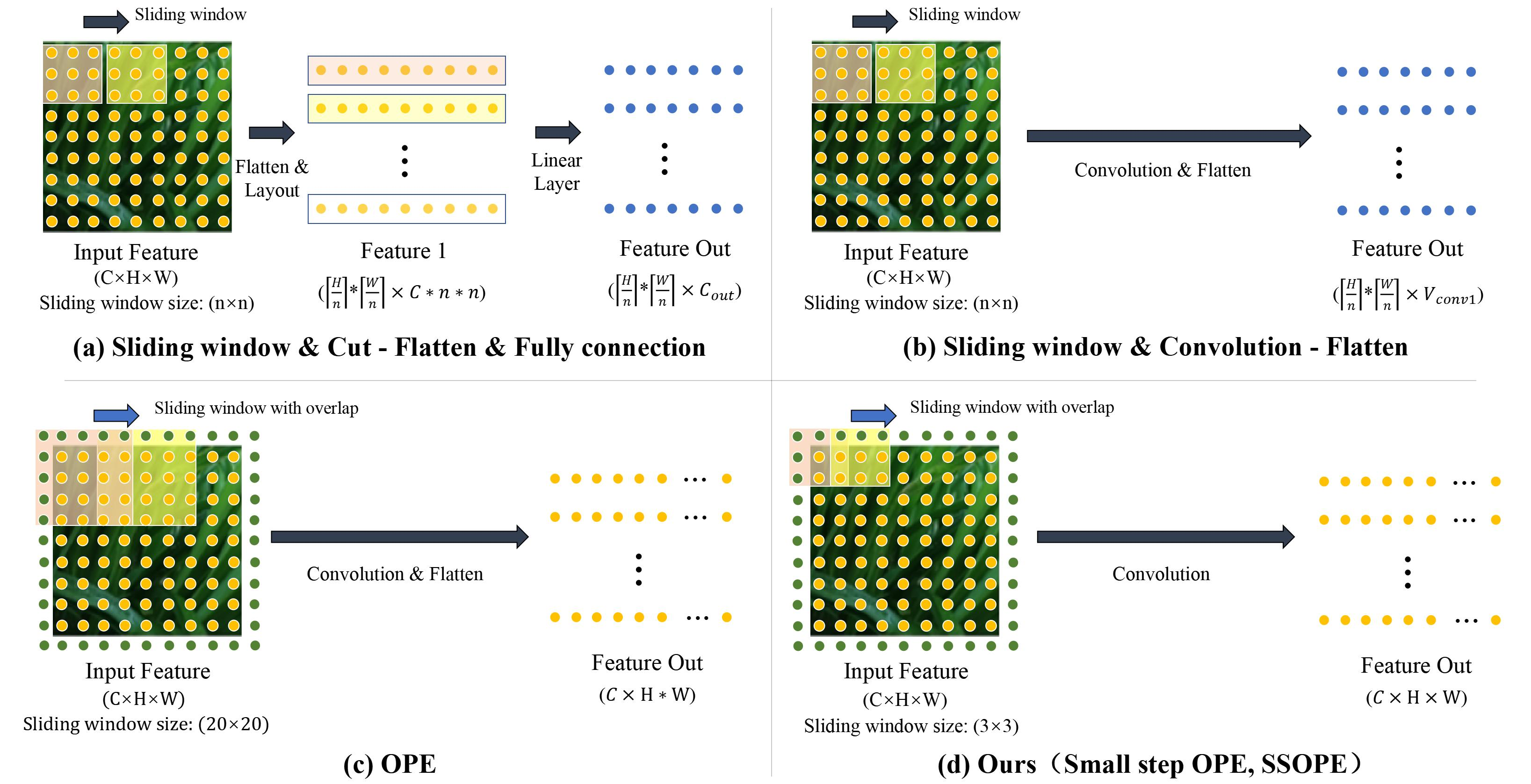}
  \caption{Patch embedding strategies: standard, OPE, and the SSOPE.}
  \label{fig:Fig2}
\end{figure}





\begin{figure}
  \centering
  \includegraphics[width=0.85\linewidth]{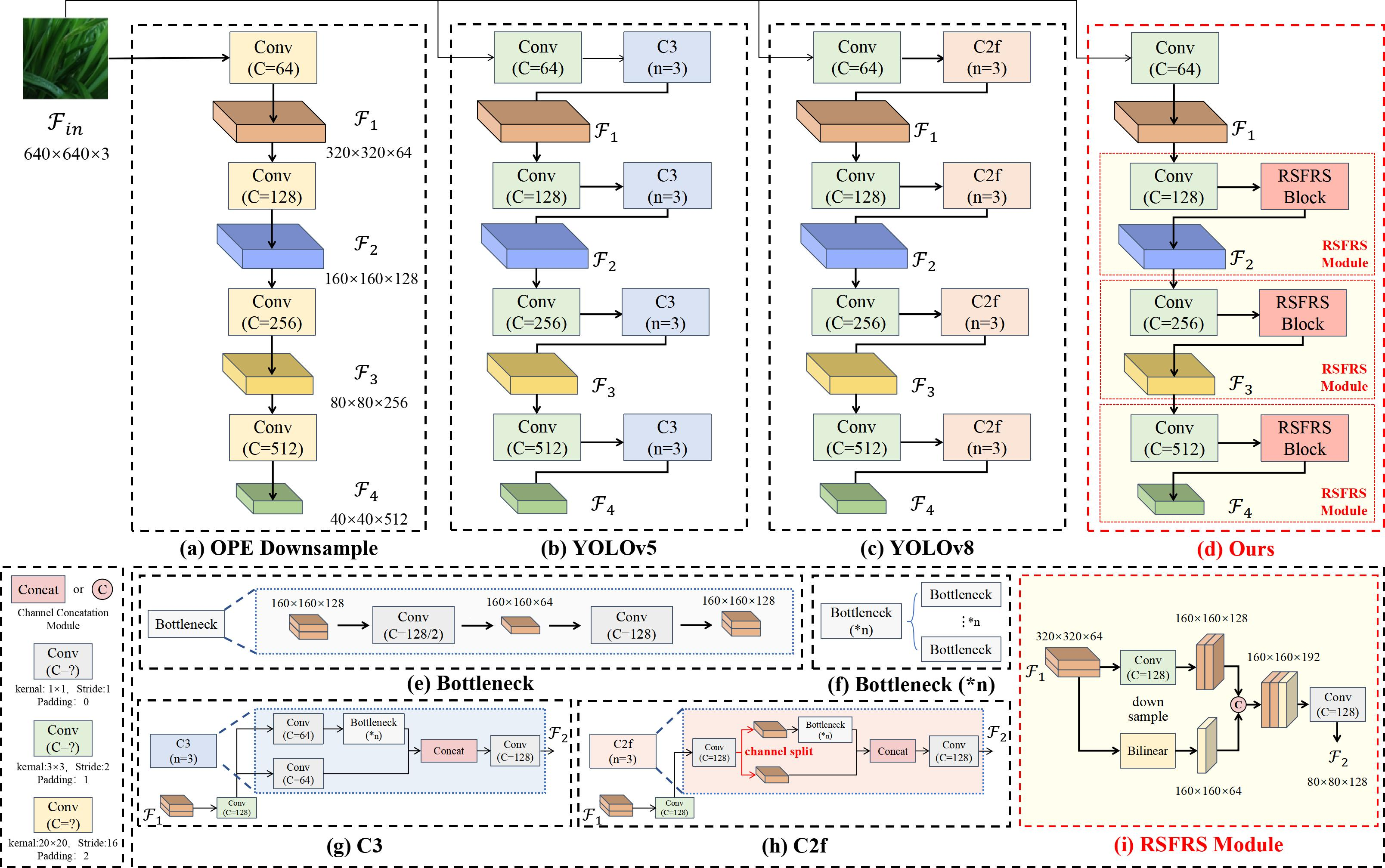}
  \caption{Comparison of downsampling modules with various channel information fusion mechanisms.}
  \label{fig:Fig3}
\end{figure}

\subsubsection{SSOPE and RSFRS}

To address the limitations of standard non-overlapping patch embeddings, which weaken spatial continuity and blur object boundaries (Fig.~\ref{fig:Fig2}(a)-(b)), we adopt a small-scale overlapping strategy. While OPE~\citep{xu2024plant} uses a $20\times20$ kernel with stride 16 to mitigate this issue (Fig.~\ref{fig:Fig2}(c)), its large receptive field risks merging foreground and background. Our SSOPE variant (Fig.~\ref{fig:Fig2}(d)) instead employs a compact $3\times3$ kernel with stride 2 to better preserve local detail. To complement this, we propose RSFRS (Fig.~\ref{fig:Fig3}(i)), which combines learnable downsampling and non-parametric bilinear sampling. Following ideas from FPN~\citep{Lin2017FPN}, RSFRS recalls high-resolution cues by fusing a $3\times3$ stride-2 convolution and bilinear interpolation via a $1\times1$ convolution:
\begin{alignat}{3}
V_2 &= \operatorname{Conv}_{3\times3}^{s=2}(V_1),  \quad
V_3 &= \operatorname{Bilinear}(V_1),  \quad
V_{\text{out}} &= \operatorname{Conv}_{1\times1}(\operatorname{Concat}(V_2, V_3)).
\end{alignat}

Together, SSOPE and RSFRS enable compact yet expressive embeddings that retain fine-grained lesion details, inspired by both OPE and the Inception design~\citep{szegedy2015going}.

\subsubsection{TCM}
Since disease symptoms primarily appear on leaf surfaces, effective lesion detection requires both local edge sensitivity and global contextual understanding. Convolutions excel at capturing high-frequency details such as leaf margins, but may mistake soil cracks or dead grass for lesions in complex field environments. In addition, scale shifts caused by varying leaf-camera distances can introduce false positives. In contrast, transformers provide a global receptive field and can model long-range spatial relationships \citep{Goyal2020Inductive}; however, their weak inductive bias may lead to missing edge details under uneven illumination.

\begin{figure}
  \centering
  \includegraphics[width=0.85\linewidth]{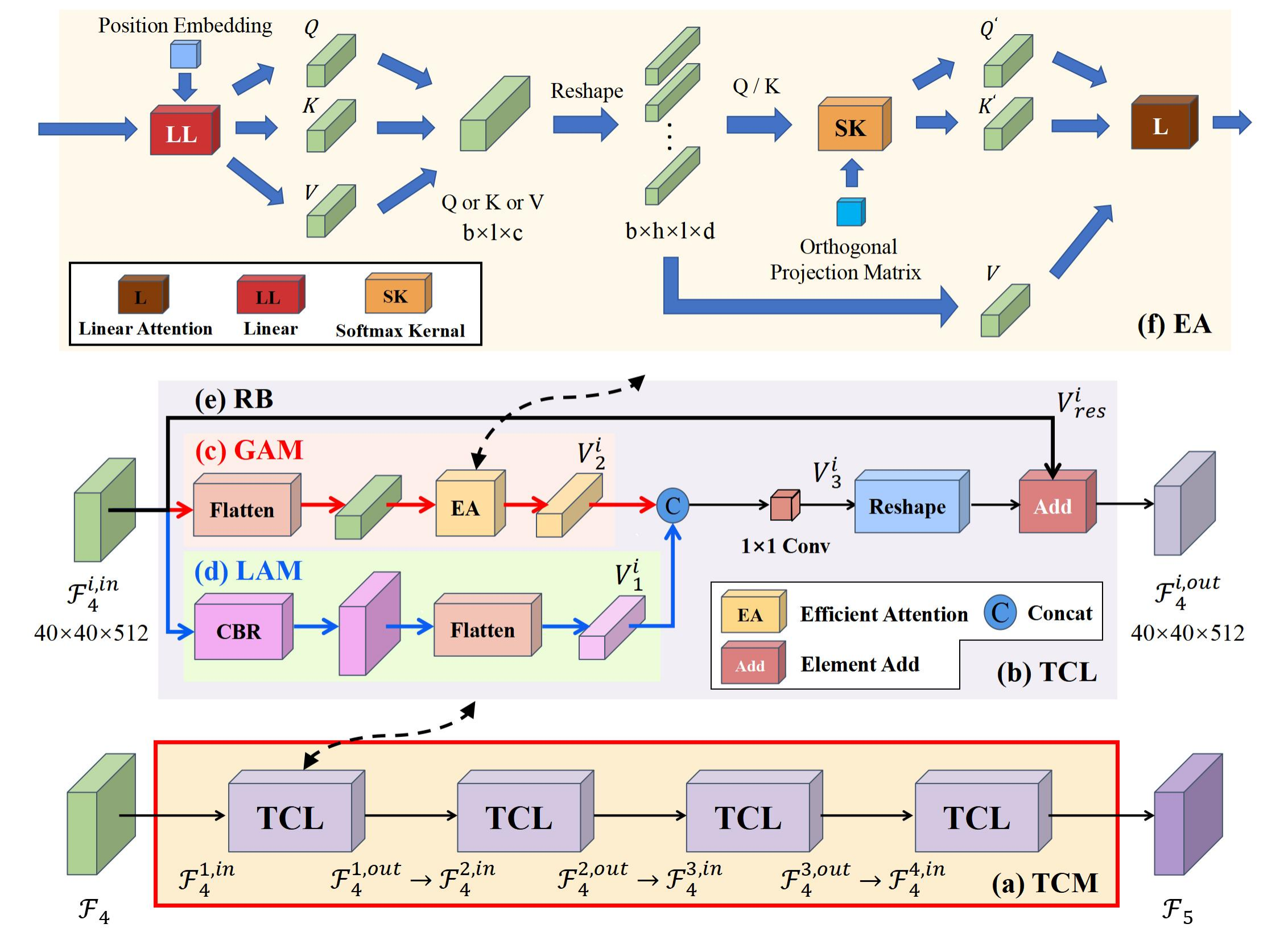}
  \caption{Structures of the TCM (a) and TCL (b) modules.}
  \label{fig:Fig4}
\end{figure}

Most existing plant-disease detectors send convolutional features through a single CNN-to-transformer stream to capture global context \citep{Thai2024MobileHTransformer, ji2021multi, 10.1007/978-3-030-87193-2_4}. Yet vanilla transformers lack strong spatial inductive bias, so these pipelines converge slowly on small agricultural datasets and often miss subtle lesion cues \citep{zhang2025depth}. We therefore introduce a parallel tri-branch transformer-convolution layer (TCL) that fuses CNN-style local attention (LAM) with transformer global attention (GAM) and a residual branch. Stacking four TCLs forms the TCM (Fig. \ref{fig:Fig4}), enabling simultaneous capture of fine-grained edges and long-range semantics.

Each TCL (Fig.~\ref{fig:Fig4}(b)) comprises three parallel branches: a global-attention module (GAM, Fig.~\ref{fig:Fig4}(c)), a local-attention module (LAM, Fig.~\ref{fig:Fig4}(d)), and a residual branch (RB, Fig.~\ref{fig:Fig4}(e)). Although this design alleviates the limitations of single branch self-attention models \citep{Thai2024MobileHTransformer}, capturing both fine lesion details and global context in complex field environments remains challenging. The GAM employs EA \citep{choromanski2020rethinking} to model long-range dependencies, the LAM uses a convolution-batch-normalization-ReLU (CBR) block to highlight local structures, and the RB preserves the original input to stabilize training.



Given an input feature map \(\mathcal{F}_4^{i,\text{in}}\), each TCL splits it into three branches. The LAM output \(V_1^i\) and GAM output \(V_2^i\) are concatenated and compressed via a \(1 \times 1\) convolution, then reshaped and added to the residual branch output \(V_{\mathrm{res}}^i\) to produce the final TCL output:
\begin{align}
\mathcal{F}_4^{i,\text{out}} = \operatorname{Reshape}\bigl(\operatorname{Conv}_{1\times1}(\operatorname{Concat}[V_1^i, V_2^i])\bigr) + V_{\mathrm{res}}^i, \quad i = 1, 2, 3, 4.
\end{align}

Repeating this process across the four stacked TCLs progressively refines both local and global representations.

\textbf{Efficient attention (EA):} 
While transformers provide essential global modeling capacity for distinguishing leaves from lesions in complex agricultural scenes, their computational cost limits practical deployment. Standard multi-head self-attention (MHSA) has quadratic complexity \(O(L^{2} d)\), where \(L = H \times W\) denotes the number of tokens extracted from a feature map of height \(H\) and width \(W\), and \(d\) is the channel (feature) dimension. This cost impedes real-time inference in resource-constrained scenarios.

To address this, we integrate EA within each TCL block. EA approximates softmax attention using randomized feature mappings, reducing complexity to linear \(O(Lmd)\) (\(m \ll L\)) while preserving global dependencies. For an input of \(8 \times 1600 \times 32\), standard MHSA requires \(\sim\)\(2.56 \times 10^{9}\) operations versus EA's \(8.19 \times 10^{5}\), a four-order magnitude reduction.  
The core EA workflow is as follows: random Gaussian matrices are sampled and processed by QR decomposition to obtain an orthogonal projection block \(Q_{\mathrm{block}} \in \mathbb{R}^{m \times d}\), which is then scaled to produce the final projection matrix \(P = \Lambda Q_{\mathrm{block}}\). Queries \(Q_i\) and keys \(K_j\) are mapped to a random feature space using
\begin{equation}
\phi(x) = \exp\left(x P^\top - \tfrac{1}{2}\|x\|^2 - c_x\right),
\end{equation}

where \(c_x\) stabilizes the computation for numerical stability. 

\subsubsection{DFPN with DAB}

\begin{figure}
  \centering
  \includegraphics[width=0.85\linewidth]{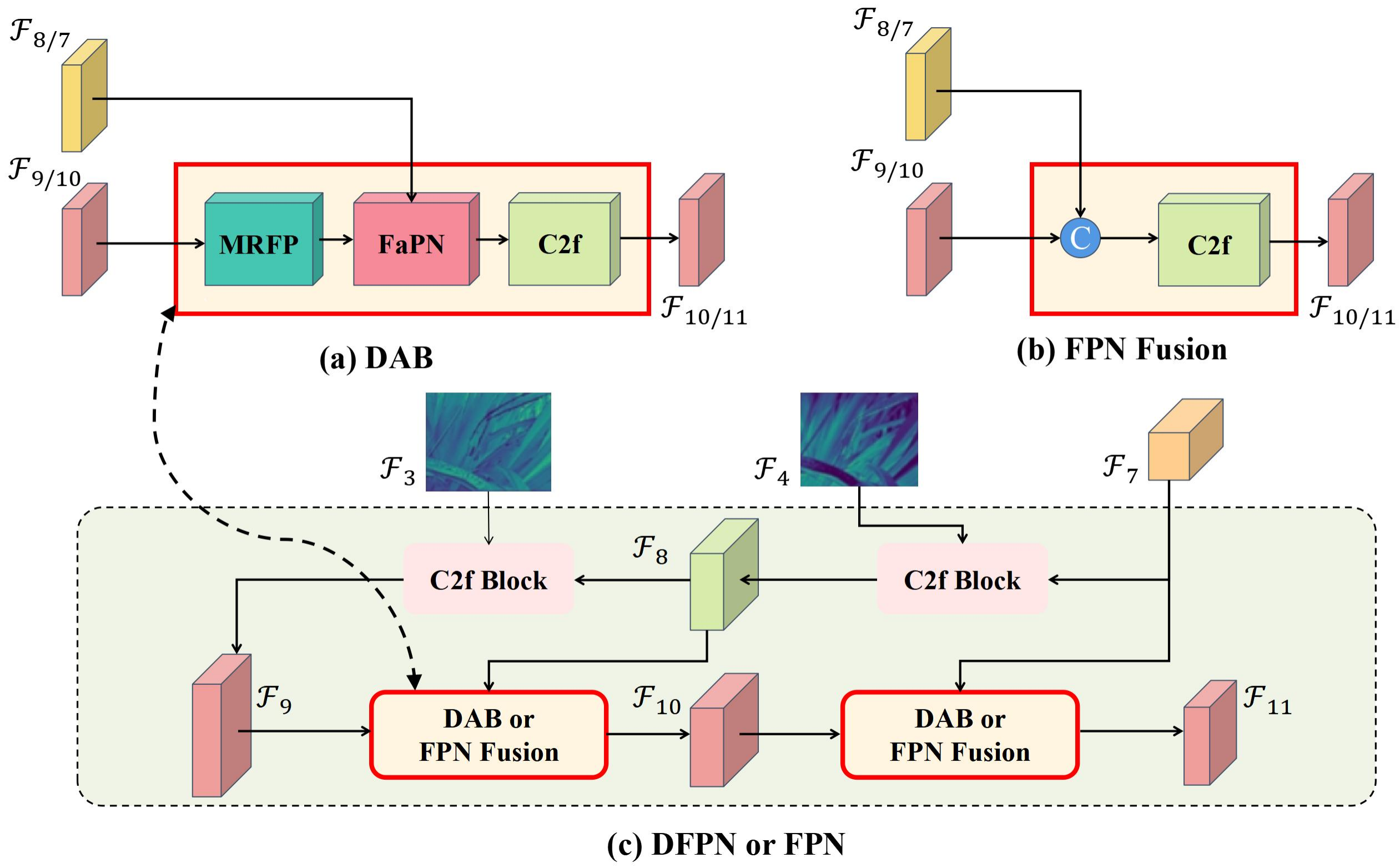}
  \caption{Architectural comparison between the proposed DFPN and the conventional FPN.}
  \label{fig:Fig5}
\end{figure}

The conventional feature pyramid network (FPN) \citep{Lin2017FPN} (Fig.~\ref{fig:Fig5}(b)) fuses semantic and spatial features through top-down pathways but suffers from semantic degradation and limited bottom-up information flow. PANet~\citep{wang2019panet} addresses this with additional bottom-up paths to enhance fine-grained details. However, \citep{huang2021fapn} shows that spatial misalignment during feature fusion, caused by downsampling, impairs localization and classification accuracy~\citep{Mingyue2022DeepLearning}.

To resolve this, we propose a deformable alignment block (DAB) that integrates three modules: MRFP, FaPN, and C2f. As shown in Fig.~\ref{fig:Fig5}, the original FPN is modified by replacing the red-boxed region in Fig.~\ref{fig:Fig5}(c) with the proposed DAB (see Fig.~\ref{fig:Fig5}(a)), resulting in the DFPN. 

In the MRFP, parallel \(5{\times}5\) and \(7{\times}7\) depthwise separable convolutions (DWConv) are employed to substantially enlarge the receptive field and aggregate multi-scale context. Previous studies have shown that the multi-receptive-field design provided by parallel, multi-scale kernels can effectively improve object localization accuracy~\citep{Liu2018RFBNet,Gao2019Res2Net}. Therefore, introducing these richer contextual features is intended to supply FaPN with more precise references for offset prediction, thereby enhancing the accuracy of the subsequent deformable alignment.

\begin{figure}
  \centering
  \includegraphics[width=0.85\linewidth]{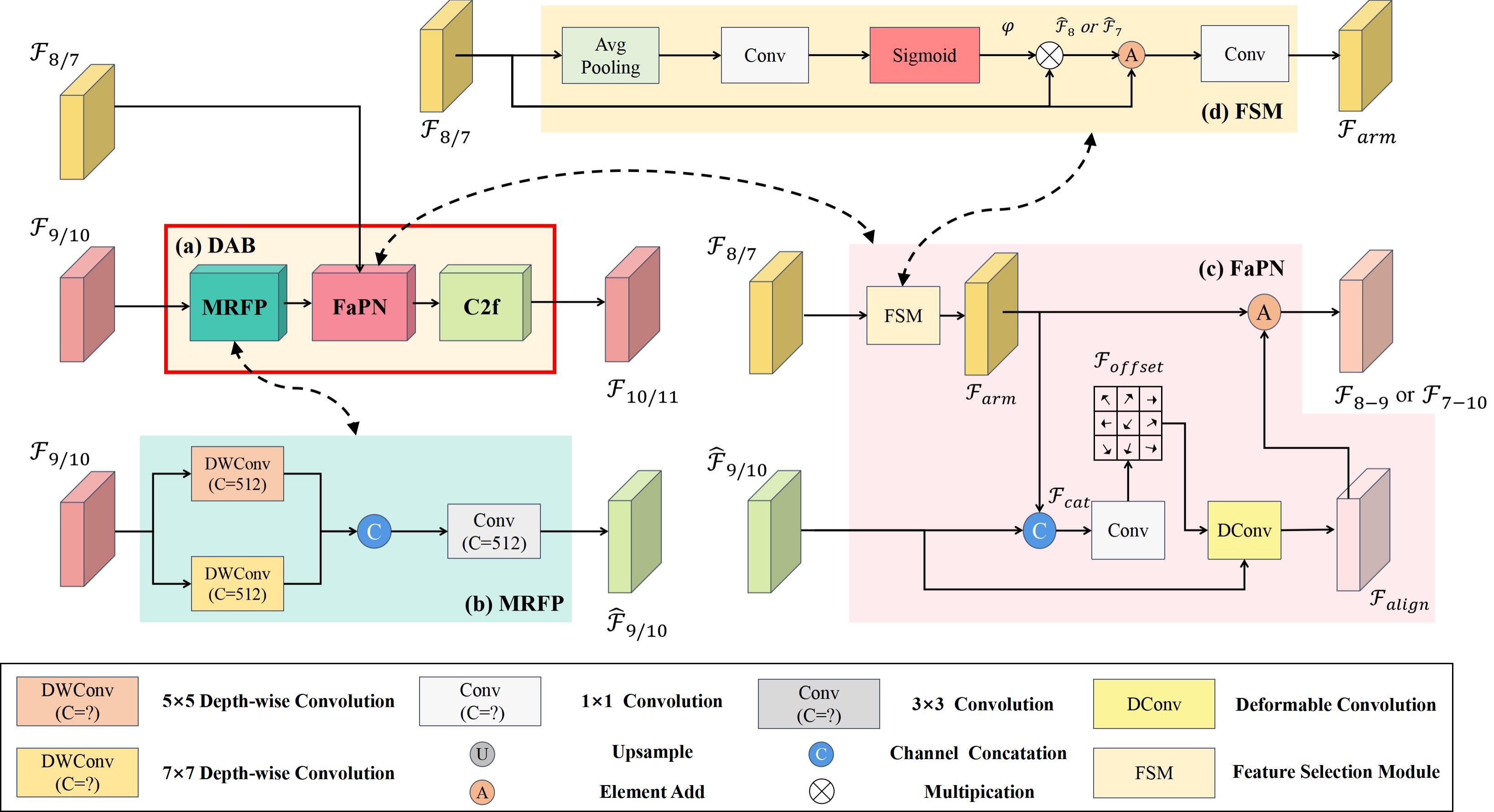}
  \caption{Architecture of the deformable alignment block (DAB).}
  \label{fig:Fig6}
\end{figure}

The architecture of DAB is detailed in Fig.~\ref{fig:Fig6}(a). The upper-level feature $F_{9/10}$ first passes through the MRFP module (Fig.~\ref{fig:Fig6}(b)), which applies $5\times5$ and $7\times7$ DWConv followed by a $1\times1$ convolution, expanding the receptive field with minimal parameter increase. The resulting feature $\hat{F}_{9/10}$ is concatenated with the corresponding lower-level feature $F_{8/7}$, and the combined tensor is fed into FaPN.

Within FaPN, the lower-level feature first enters the feature selection module (FSM), which computes channel-wise weights via global average pooling, a $1\times1$ convolution, and a sigmoid activation to produce an enhanced representation $F_{\mathrm{arm}}$. Next, $\hat{F}_{9/10}$ is concatenated with $F_{\mathrm{arm}}$ and processed by a $1\times1$ convolution to generate the offset map $F_{\text{offset}}$. This map guides a deformable convolution (DConv) that produces the aligned feature $F_{\text{align}}$. The aligned feature and $F_{\mathrm{arm}}$ are then fused (by element-wise addition), and the result is refined by a C2f block to generate the output $F_{10/11}$.

\subsection{Loss Function}

The total loss includes three terms: CIoU-based box regression loss \(L_{\text{box}}\) for precise localization, Distribution Focal Loss \(L_{\text{dfl}}\) for coordinate refinement, and binary cross-entropy \(L_{\text{cls}}\) for objectness and classification. CIoU considers overlap, distance, and aspect ratio, leading to more accurate box predictions, and the final loss:
\begin{alignat}{2}
L_{\text{box}} &= 1 - \text{CIoU}(b, b^*),  \quad
L_{\text{total}} &= \lambda_{\text{box}} L_{\text{box}} + \lambda_{\text{dfl}} L_{\text{dfl}} + \lambda_{\text{cls}} L_{\text{cls}}.
\end{alignat}

\subsection{Dataset construction}
We introduce two publicly available daylily leaf disease detection datasets annotated solely at the lesion-level, comprising 1,746 images and 7,839 lesion instances, captured under both ideal and in-field conditions.

\begin{figure}[htbp]
  \centering
  \includegraphics[width=0.85\linewidth]{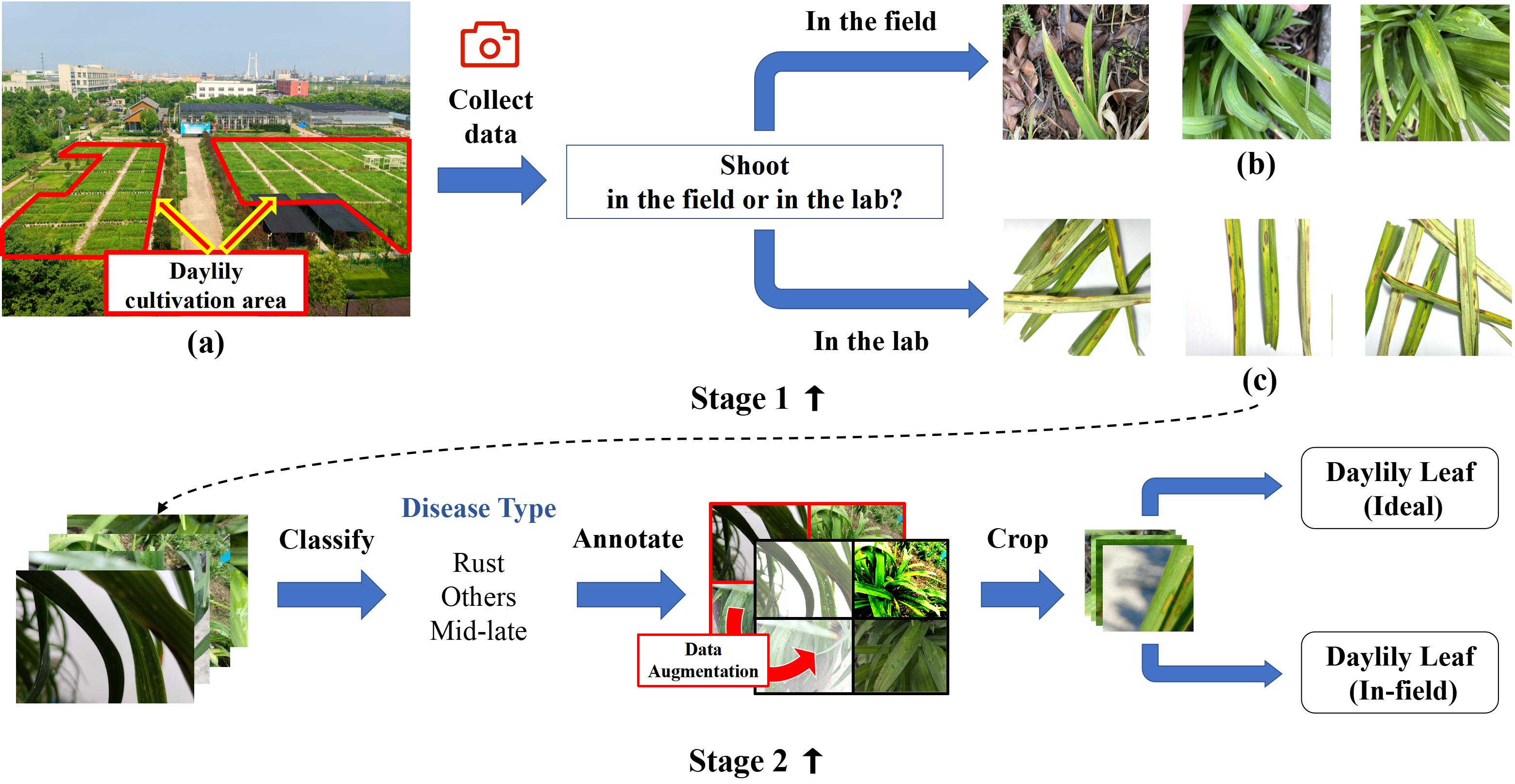}
  \caption{Overview of dataset construction. Stage 1: data acquisition, including (a) garden location, (b) in-field image samples (in-field), and (c) laboratory image samples with white backgrounds (ideal). Stage 2: data processing workflow.}
  \label{fig:Fig7}
\end{figure}

\subsubsection{Data collection}
Public leaf-image datasets are typically acquired in two settings:  field (crop-canopy) and plain-background, each supporting either classification or detection tasks~\citep{Dong2022Annotation}. Representative classification sets include PlantVillage, Turkey-PlantDataset, AES-CD9214, PlantDoc, FieldPlant, and Pigeon-pea, most of which target real-time disease detection in uncontrolled environments.

\begin{figure}[htbp]
  \centering
  \includegraphics[width=0.75\linewidth]{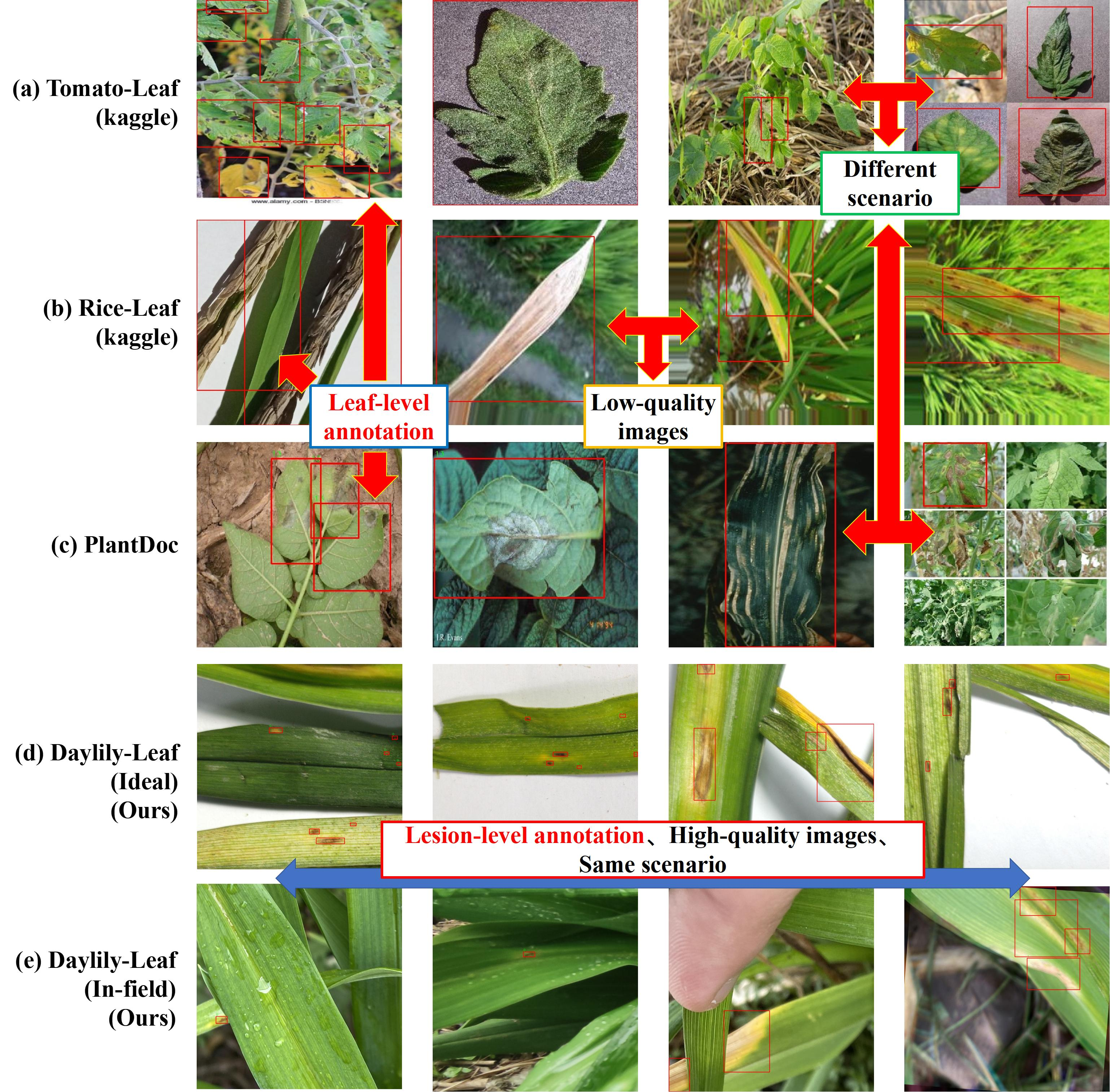}
  \caption{Comparison of datasets with manually annotated ground-truth bounding boxes (shown in red).}
  \label{fig:Fig8}
\end{figure}

Most existing public datasets for detection show diseased leaves against simple backgrounds and often provide limited image quality. For instance, the Tomato-Leaf detection set (Kaggle), the Rice-Leaf set (Kaggle), and PlantDoc (Fig.~\ref{fig:Fig8}(a)-(c)) mix field and controlled images into a single collection. The absence of high-quality, in-field data is now a major bottleneck in plant-disease detection.

To address this gap, we collected Daylily-Leaf images in Daylily cultivation area (Fig.~\ref{fig:Fig7}(a)) in two scenarios: (i) in-field scenes (Fig.~\ref{fig:Fig7}(b)) and (ii) a laboratory with a white background (Fig.~\ref{fig:Fig8}(c)). Following the pipeline in Fig.~\ref{fig:Fig7} (stage 2), we performed expert annotation, data augmentation, and sliding-window cropping to accommodate the native 12-megapixel resolution. The final Daylily-Leaf dataset is divided into ideal (Fig.~\ref{fig:Fig7}(d)) and in-field subsets (Fig.~\ref{fig:Fig8}(e)), the former for controlled evaluation and the latter for realistic deployment tests.

Compared with the tomato, rice, and PlantDoc datasets (Fig.~\ref{fig:Fig8}(a)-(c)), ours offers two advantages: (1) separate ideal and in-field subsets, enabling targeted evaluation, and (2) lesion-level (rather than leaf-level) annotations for precise localization and symptom classification.

\subsubsection{Data processing}
\begin{figure}[htbp]
  \centering
  \includegraphics[width=0.75\linewidth]{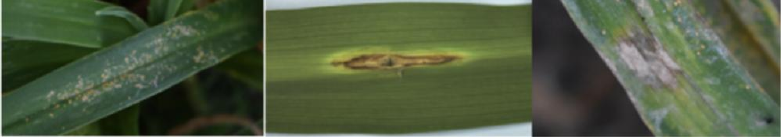}
  \caption{Biological characteristics of three major categories of plant diseases.}
  \label{fig:Fig9}
\end{figure}

\begin{figure}[htbp]
  \centering
  \includegraphics[width=0.65\linewidth]{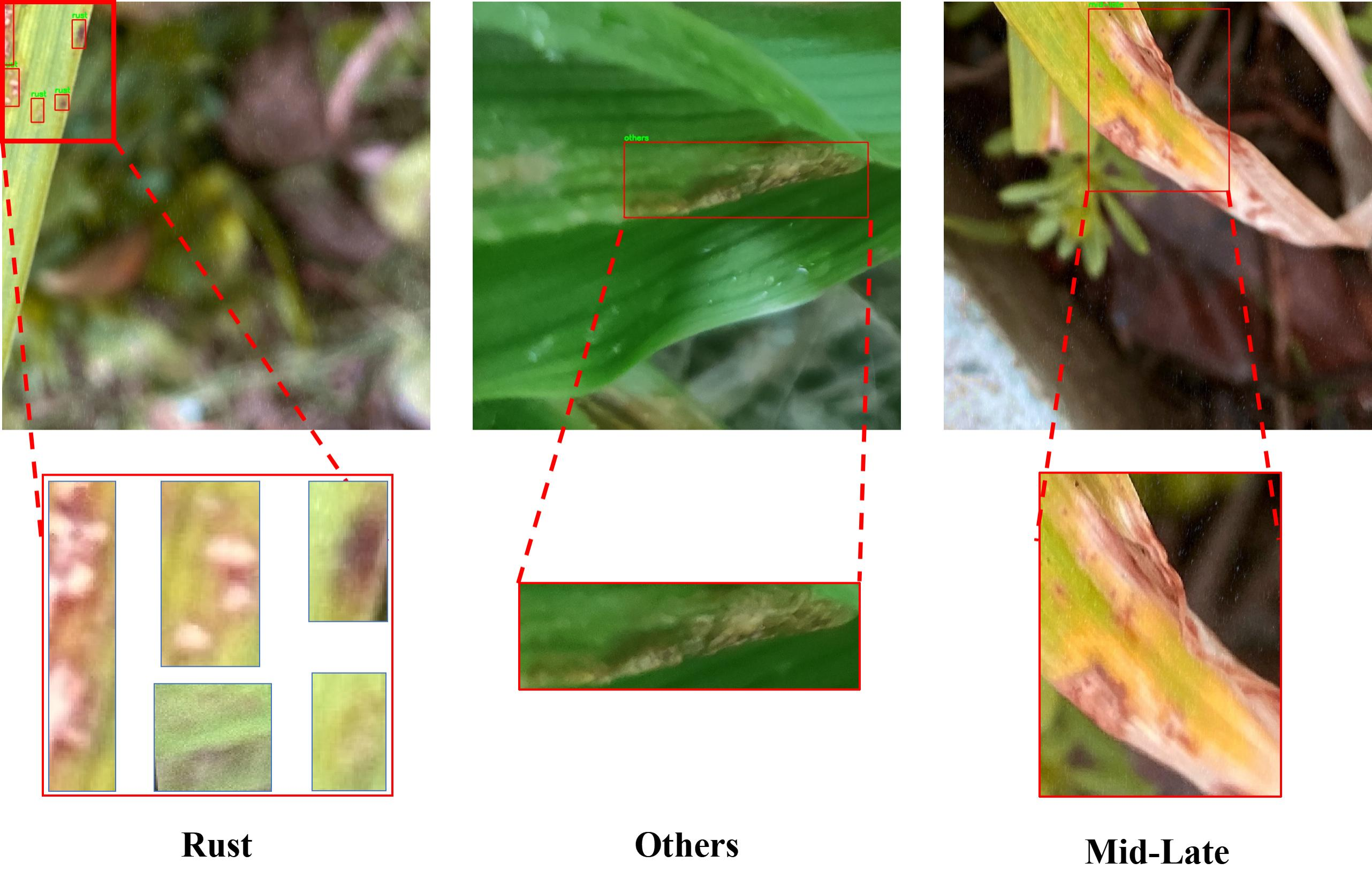}
  \caption{The three disease types defined in this study.}
  \label{fig:Fig10}
\end{figure}

Using the protocol in \citep{Dong2022Annotation}, we manually annotated fine-grained lesion regions. As shown in Fig.~\ref{fig:Fig9}, early-stage rust, leaf-spot, and powdery mildew are biologically distinct, but in some cases their visual characteristics may partially resemble each other. Moreover, due to varying real-world infection rates among disease types, the dataset exhibits inherent class imbalance. To mitigate both visual ambiguity and class imbalance, we regrouped the lesions into three broader categories, Rust, Others, and Mid-Late, based on morphological features and sample distribution. This coarse categorization simplifies the learning space and enhances model robustness in downstream detection. Although fine-grained disease labels are not the focus of this work, the current taxonomy could be extended in future studies using existing public classification datasets, enabling adaptation to more specific crop disease scenarios. All annotations were created using LabelImg, with representative examples shown in Fig.~\ref{fig:Fig10}.

To improve model generalization, we applied extensive data augmentation including flips, rotations, brightness/contrast shifts, and simulated weather effects (e.g., clouds, snow), along with saturation changes, Gaussian blur, and mosaic overlays. After processing, the Daylily-Leaf (ideal) and Daylily-Leaf (in-field) datasets jointly comprise 1,746 images and 7,839 annotated lesion instances (Table~\ref{tab:t1}), each subset further divided into training and validation sets.

\begin{table}[htbp]\rmfamily
\centering
\caption{Statistics of the Daylily-Leaf (ideal and in-field) datasets.}
\label{tab:t1}
\begin{threeparttable}
\resizebox{0.45\linewidth}{!}{%
\begin{tabular}{lllllll}
\toprule
Dataset & Subset & Img. & Tot. & R. & O. & M.L. \\
\midrule
\multirow{2}{*}{Ideal} 
    & Train & 569 & 3877 & 2229 & 1228 & 420 \\
    & Val   & 244 & 1295 &  791 &  375 & 129 \\
\midrule
\multirow{2}{*}{In-field} 
    & Train & 653 & 1788 & 1169 &  552 &  67 \\
    & Val   & 280 &  879 &  469 &  374 &  36 \\
\bottomrule
\end{tabular}
}
\vspace{1mm}

\footnotesize

\par
\raggedright
\textbf{Note.} \textbf{Img.} = Number of images;  \quad
\textbf{Tot.} = Total annotated lesions; \quad
\textbf{R.}/\textbf{O.}/\textbf{M.L.} = Numbers of rust, ``others''-type, and mid-late stage lesions, respectively.

\end{threeparttable}
\end{table}

\section{Experiments and results}
To benchmark our approach, we compare TCLeaf-Net against 13 mainstream object detectors, covering both convolution-based and transformer-based paradigms. The convolution-based baselines include YOLOv3~\citep{Redmon2018YOLOv3}, YOLOv5L~\citep{Jocher2020YOLOv5}, YOLOv6L~\citep{Li2022YOLOv6}, YOLOv8L~\citep{Ultralytics2023YOLOv8}, YOLOv9C~\citep{Zhang2024YOLOv9}, YOLOv10L~\citep{Wang2024YOLOv10}, and YOLO11L~\citep{Lee2024YOLOv11}. In addition, we include six single branch transformer-based detectors: RTDETR-R50, RTDETR-R101, RTDETR-L, and RTDETR-X~\citep{zhao2023rtdetr}, as well as the recently released YOLO12L and YOLO12X \citep{tian2025yolov12}.
All models trained from scratch and they are evaluated on both the Daylily-Leaf (ideal) and Daylily-Leaf (in-field) datasets constructed in this work, enabling a direct comparison of detection performance under controlled versus real-world agricultural conditions. Each experiment is run three times and averaged.

\subsection{Experimental setting and metrics}
During training, we employed the SGD optimizer with initial learning rate was set to 0.001 with a momentum of 0.937. The input images were resized to \(640\times640\) pixels, and trained for 200 epochs with a batch size of 16. The detailed configuration of the experimental environment is summarized in Table~\ref{tab:t2}.

\begin{table}[htbp]\rmfamily
\centering
\caption{Experimental environment.}
\label{tab:t2}
\resizebox{0.5\linewidth}{!}{%
\begin{tabular}{ll}
\toprule
Type              & Value \\ \midrule
Operating system  & Ubuntu 22.04 LTS \\
GPU               & NVIDIA RTX 4090D (24 GB) \\
CPU               & Intel Xeon Platinum 8481C \\
Software stack    & Python 3.12, PyTorch 2.5.1, CUDA 12.4 \\
\bottomrule
\end{tabular}
}
\end{table}

Model quality and efficiency are reported with the following metrics: precision~($P$), recall~($R$), F1-score~($F1$), mean average precision at 0.50 IoU~(mAP@50), mean average precision from 0.50 to 0.95 IoU~(mAP@50:95), parameter count~(Param.), and giga floating-point operations per inference~(GFLOPs). 

\begin{alignat}{3}
P  &= \frac{TP}{TP + FP},  \quad
R  &= \frac{TP}{TP + FN},  \quad
F1 &= 2 \times \frac{P \times R}{P + R}.
\end{alignat}

Higher $P$, $R$, $F1$, mAP@50, and mAP@50:95 indicate better detection performance, whereas lower Param. and GFLOPs imply a lighter and faster model. In the following tables, $\uparrow$ indicates that higher values are better, while $\downarrow$ indicates that lower values are better.

\subsection{Ablation experiment}

Table~\ref{tab:t3} summarizes the ablation results. The base model YOLOv8L (Exp.~1) attains 87.8\% and 72.8\% mAP@50 on the ideal and in-field datasets, respectively.
Introducing the TCM (Exp.~2) significantly improves precision, mAP@50, and mAP@50:95. Although recall decreases moderately, the overall F1 still increases, while also trimming the network from 43.6 M to 38.2 M parameters and reducing GFLOPs from 165.4 to 133.8.

Adding the RSFRS block (Exp.~3) increase recall on both datasets. However, precision decreases moderately on the ideal dataset, resulting in a slight dip in mAP@50 and leaving F1 almost unchanged.

Combining TCM with RSFRS (Exp.~5) or with DFPN (Exp.~6) delivers further improvements and a better balance between precision and recall.

The full model, TCLeaf-Net (Exp.~7), integrates TCM, RSFRS and DFPN, achieving the best results: 89.5\%/65.2\% mAP@50/mAP@50:95 on the ideal dataset and 78.2\%/55.1\% on the in-field dataset. Compared with the baseline, this represents gains of +1.7 pp (ideal) and +5.4 pp (in-field) in mAP@50, while reducing computational cost by 7.5 GFLOPs and saved 8.7\% GPU memory during training related to the base model.

These results demonstrate the complementary nature of the proposed modules and their effectiveness in improving detection accuracy, robustness and efficiency. Next, we further investigate the effects of individual components, beginning with detailed ablation studies on the RSFRS, SSOPE, and DFPN modules.

\begin{table*}[htbp]\rmfamily
\centering
\caption{Ablation study of each proposed components on the Daylily-Leaf (ideal and in-field) datasets.}
\label{tab:t3} 
\begin{threeparttable}
\resizebox{\textwidth}{!}{
\begin{tabular}{llllllllllllll}
\toprule
Exp. & Model 
& \multicolumn{5}{l}{Ideal dataset} 
& \multicolumn{5}{l}{In-field dataset} 
& GFLOPs $\downarrow$ & Param. (M)$\downarrow$ \\
\cmidrule(lr){3-7} \cmidrule(lr){8-12}
 &  & P (\%)$\uparrow$ & R (\%)$\uparrow$ & mAP@50 (\%)$\uparrow$ & mAP@50:95 (\%)$\uparrow$ & F1 (\%)$\uparrow$
   & P (\%)$\uparrow$ & R (\%)$\uparrow$ & mAP@50 (\%)$\uparrow$ & mAP@50:95 (\%)$\uparrow$ & F1 (\%)$\uparrow$
   &  &  \\  
\midrule
1 & Base & 82.6 & 83.5 & 87.8 & 63.1 & 83.0 & 82.4 & 65.9 & 72.8 & 50.6 & 73.2 & 165.4 & 43.6 \\
2 & Base+TCM & \textbf{86.9} & 81.1 & 88.4 & 64.0 & 83.9 & \textbf{88.8} & 63.3 & 75.4 & 54.0 & 73.9 & \textbf{133.8} & \textbf{38.2} \\
3 & Base+RSFRS & 81.6 & \textbf{83.6} & 87.5 & 63.1 & 82.6 & 82.5 & 67.5 & 74.1 & 52.2 & 74.3 & 173.8 & 46.3 \\
4 & Base+DFPN & 85.3 & 81.2 & 88.7 & 63.3 & 83.2 & 81.3 & 65.6 & 73.4 & 51.0 & 72.6 & 180.2 & 48.1 \\
5 & Base+TCM+RSFRS & 86.5 & 81.7 & 88.9 & 64.9 & 84.0 & 85.8 & 65.4 & 75.9 & 53.6 & 74.2 & 142.7 & 41.9 \\
6 & Base+TCM+DFPN & 85.2 & 82.8 & 89.1 & 65.0 & 84.0 & 85.8 & 66.8 & 76.0 & 53.0 & 75.1 & 149.0 & 42.4 \\
7 & \textbf{Base+TCM+RSFRS+DFPN (Ours)} & 85.9 & \textbf{83.6} & \textbf{89.5} & \textbf{65.2} & \textbf{84.7} & 87.2 & \textbf{69.9} & \textbf{78.2} & \textbf{55.1} & \textbf{77.6} & 157.9 & 46.1 \\

\midrule
-- & \textit{Gain (Ours vs. Base model)} & +3.3 & +0.1 & +1.7 & +2.1 & +1.7 & +4.8 & +4.0 & +5.4 & +4.5 & +4.4 & -7.5 & +2.5 \\
\bottomrule
\end{tabular}
}

\vspace{1mm}
\footnotesize

\par
\raggedright
\textbf{Note.} \textbf{Bold numbers} indicate the best results;  \quad
\textbf{Base} = Base model; \quad
\textbf{Exp.} = Experiment;  \quad
\textbf{Param.} = Parameters.
\end{threeparttable}
\end{table*}

\subsubsection{Ablation study on RSFRS and SSOPE}

Table~\ref{tab:t4} compares six patch-embedding strategies on the Daylily-Leaf (ideal and in-field) subsets.  
The non-overlapping patch embedding (PTE) with a \(16{\times}16\) window achieves only 75.4\% and 52.8\% in mAP@50 and mAP@50:95, with a recall of 63.8\%. This supports earlier reports that large, gapless patches blur edges and hinder localization \citep{touvron2022three}.


Introducing an overlapping patch embedding (OPE, \(16\times16\) kernel) elevates every evaluation metric, indicating that moderate redundancy helps retain boundary cues. When the kernel is reduced to the small stride OPE (SSOPE, \(3\times3\)), mAP@50 increases to 77.1\% and mAP@50:95 to 54.9\%, which corroborates reports that finer patches improve accuracy \citep{shu2024retina}. Conversely, enlarging the kernel to \(5\times5\) or \(7\times7\) decreases precision and recall because the broader receptive fields over-smooth small lesions, illustrating the diminishing returns of excessively large kernels \citep{li2025shiftwiseconv}.

Combining SSOPE (\(3{\times}3\)) with the RSFRS block lifts precision and recall to 87.2\% and 69.9\%, indicating that RSFRS parallel resampling recovers early information loss without increasing false positives.

\begin{table}[ht]\rmfamily
\centering
\caption{Ablation results of SSOPE and RSFRS on the Daylily-Leaf (in-field)  subset.}
\label{tab:t4}
\begin{threeparttable}
\resizebox{0.7\linewidth}{!}{%
\begin{tabular}{llllll}
\toprule
Exp. & Type & P\,(\%)$\uparrow$ & R\,(\%)$\uparrow$ & mAP@50\,(\%)$\uparrow$ & mAP@50:95\,(\%)$\uparrow$ \\
\midrule
1 & PTE                         & 84.9 & 63.8 & 75.4 & 52.8 \\
2 & OPE (\(16{\times}16\))      & \textbf{85.5} & 66.2 & 76.8 & 53.6 \\
3 & SSOPE (\(3{\times}3\))      & 85.3 & 66.7 & 77.1 & 54.9 \\
4 & SSOPE (\(5{\times}5\))      & 84.3 & 64.9 & 75.5 & 52.8 \\
5 & SSOPE (\(7{\times}7\))      & 82.8 & 65.1 & 75.8 & 53.1 \\
6 & \textbf{RSFRS (With SSOPE, ours)} & \textbf{87.2} & \textbf{69.9} & \textbf{78.2} & \textbf{55.1} \\

\bottomrule
\end{tabular}
}
\vspace{1mm}

\par

\footnotesize
\raggedright

\textbf{Note.} \textbf{PTE} = patch tokenizer with a \(16{\times}16\) kernel;  \quad
\textbf{OPE} = overlapping patch embedding;  \quad
\textbf{SSOPE (\(n{\times}n\))} = small-step OPE with \(n{\times}n\) kernel.


\end{threeparttable}
\end{table}

\subsubsection{Ablation study on DFPN}
Table~\ref{tab:t5} compares three necks on the in-field dataset. BiFPN yields 82.5\% precision, 68.6\% recall, and 75.9\% mAP@50. PANet produces 79.1\% precision, 67.4\% recall, 76.3 percent mAP@50, and 54.1 percent mAP@50:95. DFPN attains the best result. These results show that DFPN offers the most balanced performance among the three designs under identical training settings.

\begin{table}\rmfamily
\centering
\caption{Ablation results of the DFPN module on the  Daylily-Leaf (in-field) dataset.}
\label{tab:t5} 
\begin{threeparttable}

\resizebox{0.55\linewidth}{!}{%
\begin{tabular}{lllll}
\toprule
Type & P (\%)$\uparrow$ & R (\%)$\uparrow$ & mAP@50 (\%)$\uparrow$ & mAP@50:95 (\%)$\uparrow$ \\
\midrule
FPN & 85.8 & 65.4 & 75.9 & 53.6 \\
BiFPN & 82.5 & 68.6 & 75.9 & 53.1 \\
PANet & 79.1 & 67.4 & 76.3 & 54.1 \\
\textbf{DFPN (Ours)} & \textbf{87.2} & \textbf{69.9} & \textbf{78.2} & \textbf{55.1} \\

\bottomrule
\end{tabular}
}
\end{threeparttable}
\end{table}

\begin{table}[ht]\rmfamily
\centering
\caption{Comparison of model performance on Daylily-Leaf (ideal and in-field) datasets.}
\label{tab:t6} 
\begin{threeparttable}
\resizebox{\textwidth}{!}{
\begin{tabular}{llllllllllll}
\toprule
Model & \multicolumn{4}{l}{Ideal dataset} & \multicolumn{4}{l}{In-field dataset} & GFLOPs $\downarrow$ & Param. (M)$\downarrow$ \\
\cmidrule(lr){2-5} \cmidrule(lr){6-9}
 & P (\%)$\uparrow$ & R (\%)$\uparrow$ & mAP@50 (\%)$\uparrow$ & mAP@50:95 (\%)$\uparrow$
 & P (\%)$\uparrow$ & R (\%)$\uparrow$ & mAP@50 (\%)$\uparrow$ & mAP@50:95 (\%)$\uparrow$
 &  &  \\
\midrule
\textbf{Convolution-based model} \\
\midrule
YOLOv3 & 84.8 & 82.3 & \uline{88.3} & 64.2 & 78.5 & \uline{68.1} & \uline{74.2} & \uline{52.6} & 282.2 & 103.7 \\
YOLOv5L & 80.7 & 82.4 & 86.8 & 64.1 & 80.9 & 64.3 & 70.0 & 47.9 & 134.7 & 53.1 \\
YOLOv6L & 81.8 & 75.9 & 84.1 & 57.4 & 37.3 & 34.4 & 27.8 & 14.1 & 391.2 & 110.9 \\
YOLOv8L & 82.6 & \uline{83.5} & 87.8 & \uline{63.1} & \uline{82.4} & 65.9 & 72.8 & 50.6 & 164.8 & 43.6 \\

YOLOv9C & \uline{85.7} & 79.4 & 87.5 & 61.2 & 80.8 & 63.0 & 72.1 & 48.9 & 102.3 & 25.3 \\
YOLOv10L & 73.5 & 68.1 & 76.1 & 58.1 & 58.5 & 49.8 & 53.5 & 34.5 & 126.3 & 25.7 \\
YOLO11L & 80.2 & 76.9 & 84.9 & 58.1 & 66.2 & 58.6 & 63.4 & 40.3 & \uline{86.6} & 25.3 \\
\midrule
\textbf{Single branch transformer-based model} \\
\midrule
RTDETR-L    & 63.0 & 62.4 & 61.9 & 39.1 & 53.5 & 49.0 & 47.7 & 28.7 & 108.0 & 32.8 \\
RTDETR-X    & 55.6 & 56.0 & 54.1 & 33.1 & 43.8 & 48.5 & 38.9 & 23.0 & 232.4 & 67.3 \\
RTDETR-R50  & 67.1 & 67.0 & 67.6 & 43.0 & 57.8 & 53.4 & 52.6 & 32.6 & 130.5 & 42.8 \\
RTDETR-R101 & 63.2 & 65.1 & 65.4 & 42.3 & 58.2 & 52.7 & 52.6 & 33.1 & 191.4 & 61.8 \\
YOLO12X    & 81.6 & 81.5 & 86.4 & 60.9 & 75.0 & 62.7 & 69.2 & 46.3 & 199.9 & 59.1 \\
YOLO12L    & 81.6 & 78.4 & 84.8 & 58.3 & 80.5 & 59.7 & 68.7 & 45.2 & 89.4  & 26.4 \\

\midrule
\textbf{Transformer-convolution hybrid model} \\
\midrule

\textbf{TCLeaf-Net (ours)} & \textbf{85.9} & \textbf{83.6} & \textbf{89.5} & \textbf{65.2} & \textbf{87.2} & \textbf{69.9} & \textbf{78.2} & \textbf{55.1} & 157.9 & 46.1 \\

\midrule

\textit{Gain (Ours vs. 2nd best)} & +0.2 & +0.1 & +1.2 & +1.0 & +4.8 & +1.8 & +4.0 & +2.5 & -- & -- \\
\bottomrule
\end{tabular}
}
\vspace{1mm}
\begin{tablenotes}[flushleft]\footnotesize
\item \textbf{Note.} \textbf{Bold numbers} indicate the best result; \uline{underlining} marks the second (2nd) best;  \quad
\textbf{OOM} = out-of-memory error (GPU).
\end{tablenotes}
\end{threeparttable}
\end{table}

\subsection{Comparative experiment}
\begin{figure}
  \centering
  \includegraphics[width=0.85\linewidth]{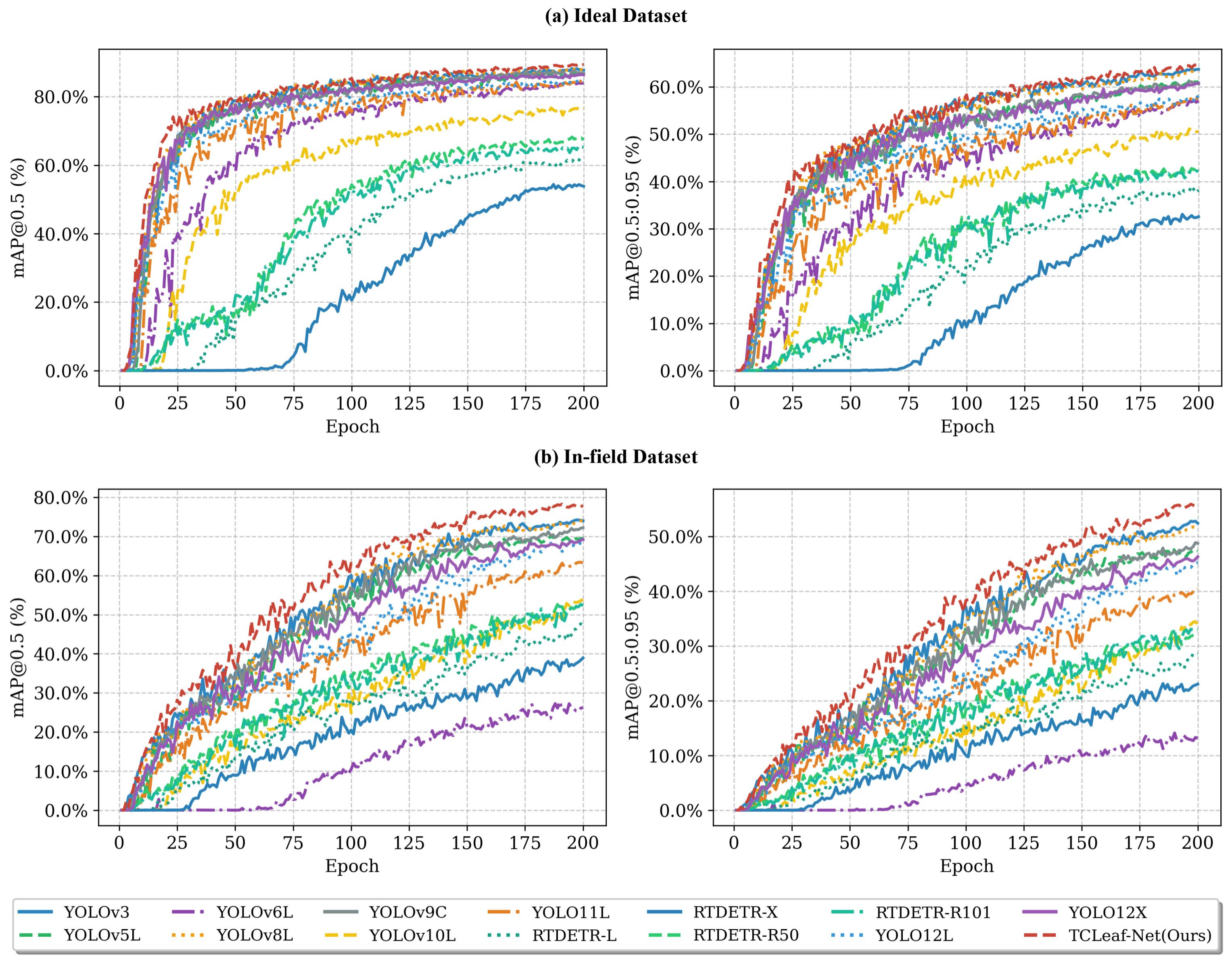}
  \caption{Comparative experiments of TCLeaf-Net and other models (epoch-mAP@50 curves): (a) trained on the  Daylily-Leaf (in-field) dataset; (b) trained on the  Daylily-Leaf (ideal) dataset.}
  \label{fig:Fig11}
\end{figure}

Table~\ref{tab:t6} compares eight detectors on the Daylily-Leaf (ideal) and Daylily-Leaf (in-field) sets using mAP@50, mAP@50:95, precision and recall, while the training curves appear in Fig.~\ref{fig:Fig11}.

As Fig.~\ref{fig:Fig11}(a)-(b) show, most of the models reach at least 60\% mAP@50 when trained and validated on the ideal dataset. When trained on the in-field dataset, however, their validation scores drop; YOLOv6L, YOLOv10L, YOLO11L and YOLO12X decline the most, revealing limited robustness to complex backgrounds, whereas the remaining models remain relatively stable.

\begin{figure}
  \centering
  \includegraphics[width=0.9\linewidth]{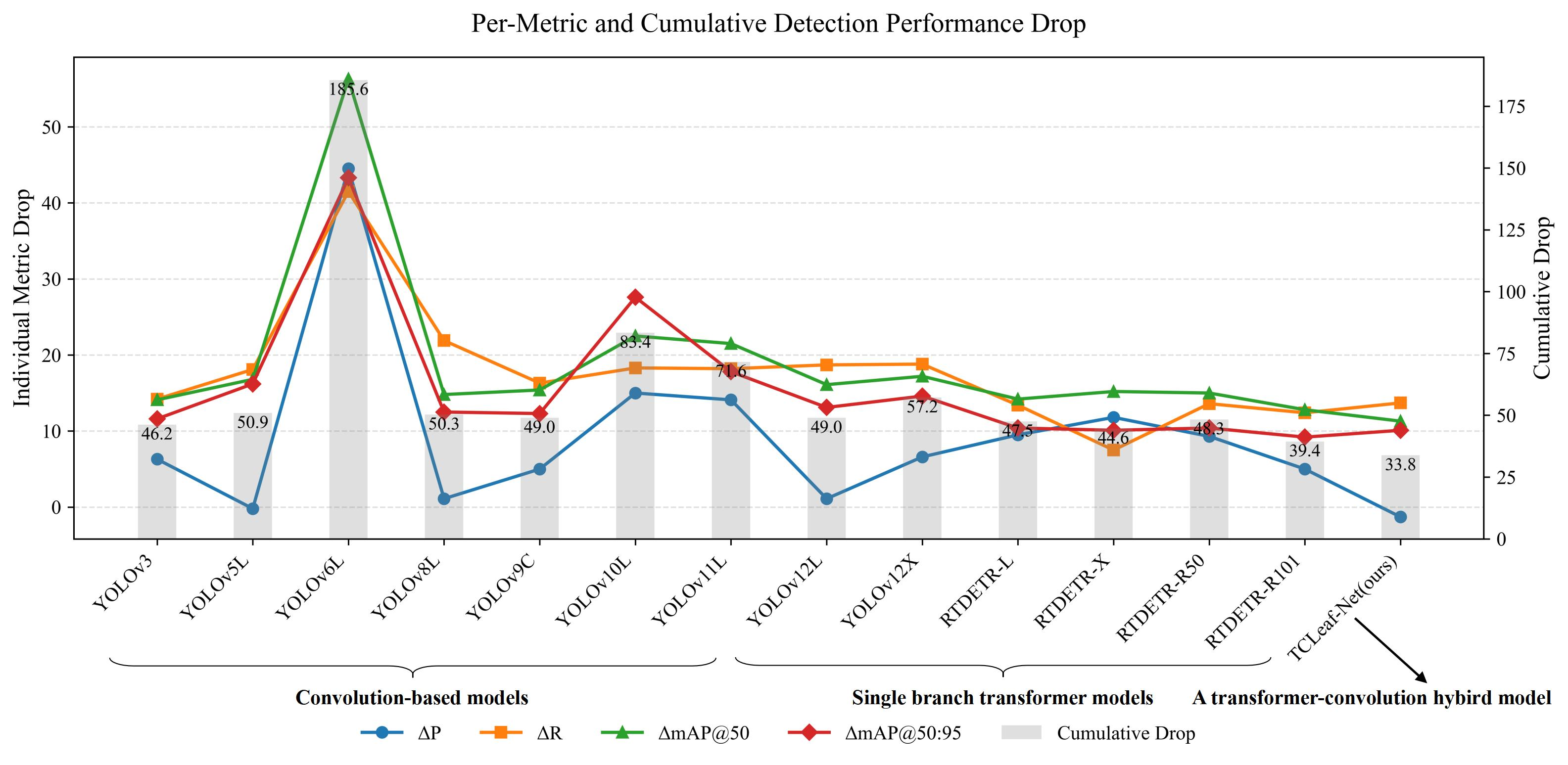}
  \caption{Performance drop from  Daylily-Leaf (ideal) to (in-field); values are in percentages (lower is better).}
  \label{fig:Fig12}
\end{figure}

TCLeaf-Net delivers the best overall performance on the in-field dataset. Although YOLOv3 ranks second-best overall performance, it requires approximately \(1.79\times\) more computation (GFLOPs) and \(2.25\times\) more parameters than TCLeaf-Net. These efficiency gains stem from the TC backbone (Table~\ref{tab:t3}, Exp. 2 and 5), whose TCM and RSFRS modules shorten the depth and reduce use of C2f blocks, cutting approximately GFLOPs by 13.7\% and parameters by 3.9\% relative to the base model YOLOv8L. By combining long-range spatial modeling with local feature attention, the TC backbone better captures relationships among leaves, lesions and background, which in turn lifts both precision and recall. On the ideal dataset, TCLeaf-Net also generalizes well, attaining 89.5\% mAP@50, 65.2\% mAP@50:95, 85.9\% precision and 83.6\% recall, again topping every competitor.

To gauge robustness, we computed the drop from the ideal to the in-field dataset across four metrics for each detector:

To gauge model robustness, we quantified the performance drop from the ideal to the in-field dataset across four key metrics-precision ($\Delta P$), recall ($\Delta R$), mAP@50 ($\Delta \text{mAP@50}$), and mAP@50:95 ($\Delta \text{mAP@50:95}$), for each detector:
\begin{align}
\Delta\text{Metric}= \text{Metric}_{\text{ideal}} - \text{Metric}_{\text{in-field}}.
\end{align}

We summed these metric differences to calculate an overall sensitivity measure (cumulative performance drop). As shown in Fig.~\ref{fig:Fig12}, TCLeaf-Net achieves the smallest cumulative performance drop (33.8 percentage points, pp), indicating superior robustness to variations in background complexity. RTDETR-R101 follows closely with a cumulative drop of 39.4 pp, while YOLO-series models generally exhibit higher sensitivity (ranging from 46.2 to 83.4 pp). Notably, YOLOv6L shows the greatest performance degradation (185.6 pp), highlighting its vulnerability to challenging in-field conditions.

Interestingly, TCLeaf-Net demonstrates a negative precision drop ($\Delta P = -1.3$ pp), suggesting improved precision in real-world scenarios, but the recall dropped. RT-DETR variants consistently exhibit moderate drops in recall but maintain relatively stable mAP scores, reflecting balanced robustness. In contrast, most YOLO models display larger and less consistent performance reductions across all metrics. Overall, TCLeaf-Net provides the best trade-off among all models, underscoring its stability and reliability in complex, practical agricultural environments.

To provide qualitative insights into how TCLeaf-Net achieves improved performance, we visualize representative detection scenarios from the in-field validation set.

\subsection{Visualization}
\begin{figure}[htbp]
  \centering
  \includegraphics[width=0.85\linewidth]{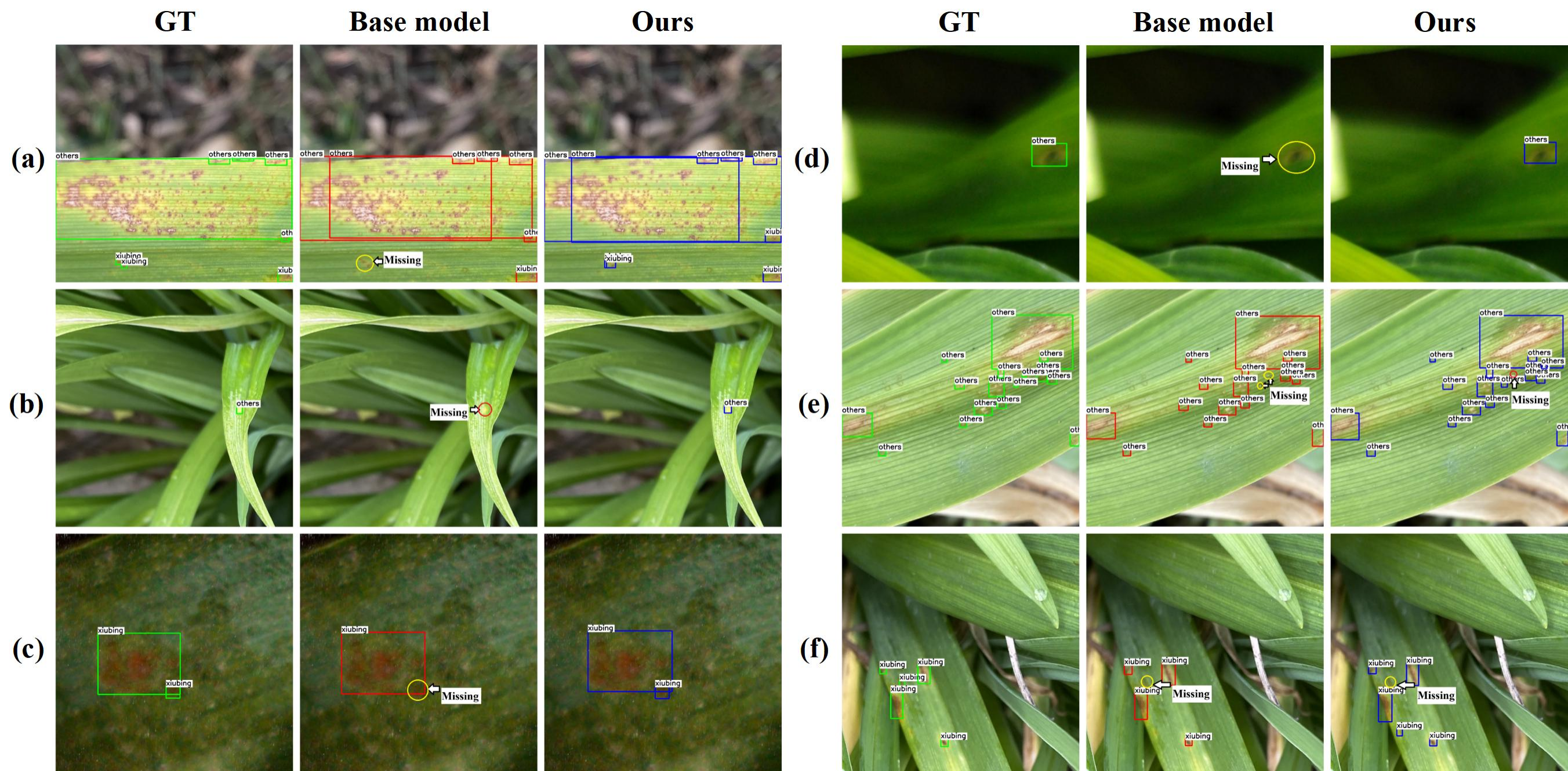}
  \caption{Detection results: ours vs. base model.}
  \label{fig:Fig13}
\end{figure}

For a qualitative assessment under real-world conditions, six representative scenarios from the in-field validation set were chosen (Fig.~\ref{fig:Fig13}(a)-(f)), illustrating the following challenges: (a) well-lit with leaves and background each occupying half the frame, (b) well-lit with multiple leaves, (c) low light with dense foliage and subtle, small lesions, (d) low light with multiple leaves and de-focused lesion areas, (e) well-lit, leaf-dominated scenes containing small lesion targets, and (f) well-lit with multiple leaves.

Fig.~\ref{fig:Fig13} presents a side-by-side comparison of the ground truth (GT; left), defined as images with human annotations, the output of the base model  (middle), and that of TCLeaf-Net (right). False positives in the model predictions are circled in white, and missed lesions are labeled as “missing”.

In detail, (a) the base model missed a partially occluded rust lesion; (b) and (c) exhibited confusion between lesion and non-lesion regions, resulting in missed detections; (d) defocused areas caused detection failure for the base model; (e) the base model failed to detect two small lesions, while TCLeaf-Net missed only one; and (f) TCLeaf-Net identified a lesion absent from the ground truth that the base model also failed to detect.

These examples show that TCLeaf-Net demonstrates greater stability in the presence of complex backgrounds and blurring. While TCLeaf-Net shows superior performance on our own Daylily-Leaf datasets, it is essential to confirm whether these improvements generalize to broader scenarios. Thus, we evaluate our model on three external, publicly available datasets.

\subsection{Generalization experiments}{}

To assess the generalization of TCLeaf-Net beyond Daylily-Leaf datasets, we evaluated its performance against multiple YOLO and RT-DETR baselines on three widely used public leaf-disease detection datasets: PlantDoc, Tomato-Leaf, and Rice-Leaf, which represent varying degrees of scene complexity and image quality.

\textbf{PlantDoc.}
PlantDoc comprises in-field images characterized by complex backgrounds and notable class imbalance. As reported in Table~\ref{tab:t7}, TCLeaf-Net achieves the highest mAP@50 (65.7\%) and mAP@50:95 (52.5\%), outperforming the strongest baseline (YOLO12X) by 2.1 and 2.7 percentage points, respectively. Although YOLO12X achieves slightly higher recall, TCLeaf-Net provides a superior balance between precision and recall, as indicated by its leading F1 (60.8\%). In addition, TCLeaf-Net attains these results with fewer parameters and lower computational cost compared to YOLO12X (46.1M vs. 59.1M parameters; 157.9 vs. 191.4 GFLOPs), further demonstrating its efficiency and practical value. These results highlight the effectiveness of TCLeaf-Net in challenging, real-world environments without reliance on large-scale pre-training.

\textbf{Tomato-Leaf and Rice-Leaf.}
On the Tomato-Leaf dataset (Table~\ref{tab:t8} (left)), which contains relatively clean and controlled images, TCLeaf-Net attains the highest recall (88.8\%) and mAP@50:95 (86.7\%), while matching the best mAP@50 (94.6\%). Its F1 (90.3\%) also ranks highest, indicating robust performance even on simpler data. For the Rice-Leaf dataset (Table~\ref{tab:t8} (right)), which features lower image quality and increased visual noise, TCLeaf-Net again secures the best mAP@50 (57.3\%) and mAP@50:95 (34.9\%), as well as the top precision (67.8\%) and a shared best F1 (59.9\%).

Overall, TCLeaf-Net consistently achieves the best results or matches the best baseline across all three benchmarks, demonstrating strong cross-dataset generalization and resistance to overfitting. These findings confirm that the model's improvements are not confined to daylily imagery but generalize to diverse plant disease detection scenarios.

\begin{table}[htbp]\rmfamily
\caption{Performance comparison on the PlantDoc dataset.}
\label{tab:t7}
\centering
\renewcommand{\arraystretch}{1.25}
\setlength{\tabcolsep}{5pt}
\footnotesize
\resizebox{0.7\linewidth}{!}{%
\begin{tabular}{llllllll}
\toprule
Model & PT & P(\%)$\uparrow$ & R(\%)$\uparrow$ & mAP@50(\%)$\uparrow$ & mAP@50:95(\%)$\uparrow$ & F1(\%)$\uparrow$ \\
\midrule
YOLOv3\citep{shill2021plant}        & COCO & 53.0 & 57.0 & 53.1 & --   & 55.0 \\
YOLOv4\citep{shill2021plant}        & COCO & 53.0 & 62.0 & 55.5 & --   & 56.0 \\
NanoDet-Plus\citep{li2022improved}  & No M. & --   & 54.2 & 55.3 & --   & --   \\
Improved-YOLOv5\citep{li2022improved}& No M. & --   & 55.0 & 58.2 & --   & --   \\
YOLOv8L\citep{moupojou2023fieldplant}& COCO & --   & --   & 55.0 & --   & --   \\
YOLOv6L\citep{Li2022YOLOv6}         & None & 25.5 & 49.6 & 39.0 & 30.8 & 33.7 \\
YOLOv9C\citep{Zhang2024YOLOv9}      & None & 57.4 & 58.9 & 60.9 & 47.9 & 58.1 \\
YOLOv10L\citep{Wang2024YOLOv10}     & None & 47.7 & 56.5 & 55.5 & 44.3 & 51.8 \\
YOLO11L\citep{Lee2024YOLOv11}      & None & 52.2 & 53.5 & 57.2 & 45.8 & 52.8 \\
YOLO12L\citep{tian2025yolov12}     & None & 58.2 & 56.9 & 61.8 & 48.9 & 57.5 \\
YOLO12X\citep{tian2025yolov12}     & None & 53.3 & \textbf{65.3} & 63.6 & 49.8 & 58.6 \\
RTDETR-L\citep{zhao2023rtdetr}      & None & 55.6 & 49.6 & 46.4 & 36.4 & 52.5 \\
RTDETR-X\citep{zhao2023rtdetr}      & None & 52.3 & 50.1 & 47.6 & 37.6 & 51.2 \\
RTDETR-R50\citep{zhao2023rtdetr}    & None & 53.2 & 54.6 & 50.6 & 40.1 & 53.9 \\
RTDETR-R101\citep{zhao2023rtdetr}   & None & 56.1 & 53.3 & 51.9 & 40.6 & 54.7 \\
\textbf{TCLeaf-Net}                 & None & \textbf{59.4}   & 62.2   & \textbf{65.7} & \textbf{52.5} & \textbf{60.8}   \\
\bottomrule
\end{tabular}
}
\vspace{2pt}
\footnotesize

\par
\raggedright
\textbf{Note.} 
\textbf{PT}: Pretrained weights used. \quad  
\textbf{No M.}: No mention. \quad  
\textbf{COCO}: COCO pretrained. \quad  
\textbf{None}: Trained from scratch. \\
The symbol "\textbf{“--”}" indicates that the metric is not reported in the cited work.
\end{table}

\begin{table*}[ht]\rmfamily
\centering
\caption{Comparison of model performance on Tomato-Leaf and Rice-Leaf datasets.}
\label{tab:t8} 
\begin{threeparttable}
\resizebox{\textwidth}{!}{
\begin{tabular}{llllllllllll}
\toprule
Model & \multicolumn{5}{l}{Tomato-Leaf} & \multicolumn{5}{l}{Rice-Leaf}  \\
\cmidrule(lr){2-6} \cmidrule(lr){7-11}
 & P (\%)$\uparrow$ & R (\%)$\uparrow$ & mAP@50 (\%)$\uparrow$ & mAP@50:95 (\%)$\uparrow$ & F1 (\%)$\uparrow$
 & P (\%)$\uparrow$ & R (\%)$\uparrow$ & mAP@50 (\%)$\uparrow$ & mAP@50:95 (\%)$\uparrow$ & F1 (\%)$\uparrow$ \\
\midrule
YOLOv3L      & \textbf{92.6} & 87.6 & 94.4 & 85.5 & 90.1 &
               63.2 & 52.1 & 53.6 & 33.1 & 57.2 \\
YOLOv5L      & 88.5 & 79.7 & 90.0 & 80.4 & 83.9 &
               64.8 & 55.0 & 56.0 & 32.6 & 59.5 \\
YOLOv6L      & \textbf{92.6} & 87.1 & 94.4 & 85.9 & 89.8 &
               50.8 & 46.4 & 46.4 & 26.8 & 48.5 \\
YOLOv8L      & 90.9 & 88.6 & \textbf{94.6} & 85.8 & 89.7 &
               65.9 & 54.9 & 56.6 & 34.6 & \textbf{59.9} \\
YOLOv9C      & 85.3 & 79.5 & 89.0 & 77.4 & 82.3 &
               64.7 & \textbf{55.3} & 56.1 & 33.8 & 59.6 \\
YOLOv10L     & 91.9 & 85.8 & 94.2 & 84.7 & 88.7 &
               53.9 & 48.6 & 46.6 & 26.7 & 51.1 \\
YOLO11L     & 91.0 & 88.7 & \textbf{94.6} & 85.8 & 89.9 &
               60.2 & 52.6 & 54.0 & 33.3 & 56.2 \\
YOLO12L     & 92.1 & 87.0 & 94.3 & 85.6 & 89.5 &
               61.3 & 54.6 & 55.9 & 34.0 & 57.8 \\
YOLO12X     & 91.9 & 87.9 & 94.3 & 85.3 & 89.8 &
               61.9 & 54.5 & 57.0 & 34.3 & 58.0 \\
RTDETR-L     & 87.5 & 80.8 & 88.5 & 78.6 & 84.0 &
               52.1 & 50.0 & 46.3 & 28.8 & 51.1 \\
RTDETR-X     & 87.2 & 80.8 & 88.5 & 78.9 & 83.9 &
               55.0 & 44.7 & 44.8 & 27.8 & 49.3 \\
RTDETR-R50   & 89.8 & 83.6 & 90.3 & 81.7 & 86.6 &
               52.7 & 44.6 & 45.2 & 27.6 & 48.3 \\
RTDETR-R101  & 88.7 & 84.0 & 90.4 & 81.9 & 86.3 &
               55.3 & 47.1 & 47.0 & 28.1 & 50.9 \\
\textbf{TCLeaf-Net} &
91.8 & \textbf{88.8} & \textbf{94.6} & \textbf{86.7} & \textbf{90.3} &
\textbf{67.8} & 53.6 & \textbf{57.3} & \textbf{34.9} & \textbf{59.9} \\
\bottomrule
\end{tabular}
}
\vspace{1mm}
\begin{tablenotes}[flushleft]\footnotesize
\item \textbf{Note.} All models trained from scratch.
\end{tablenotes}
\end{threeparttable}
\end{table*}

\section{Discussion}
This study addresses two long-standing issues in plant-disease detection: (i) the scarcity of high-quality, field-collected public datasets and (ii) the poor robustness of existing detectors in complex outdoor scenes. We therefore publish two lesion-level, daylily leaf datasets and propose TCLeaf-Net, a detector that remains reliable under real-world conditions. TCLeaf-Net is built from three interchangeable blocks: TCM, RSFRS and DFPN. These blocks are responsible for spatial modeling, semantic recovery and multi-scale fusion, respectively. This design is transferable to other crops.

Ablation studies (Table~\ref{tab:t3} (In-field dataset)) confirm that every block improves performance, with TCM contributing the largest single gain. RSFRS alone yields notable improvements, particularly on recall and overall mAP, and its combination with TCM produces a marked additional boost, indicating effective semantic recall. Adding DFPN further raises recall and F1, although a slight drop in mAP@50:95 on the in-field dataset reveals a minor trade-off in localization precision. Future work will tune deformable-convolution settings and cross-resolution fusion within DFPN to mitigate this effect.

To probe how TCLeaf-Net makes its decisions, we visualized Grad-CAM heatmaps~\citep{Selvaraju2017GradCAM} generated by its three detection heads. The maps reveal which image regions influence each head and thus guide further design refinements.

\begin{figure}[htbp]
  \centering
  \includegraphics[width=0.85\linewidth]{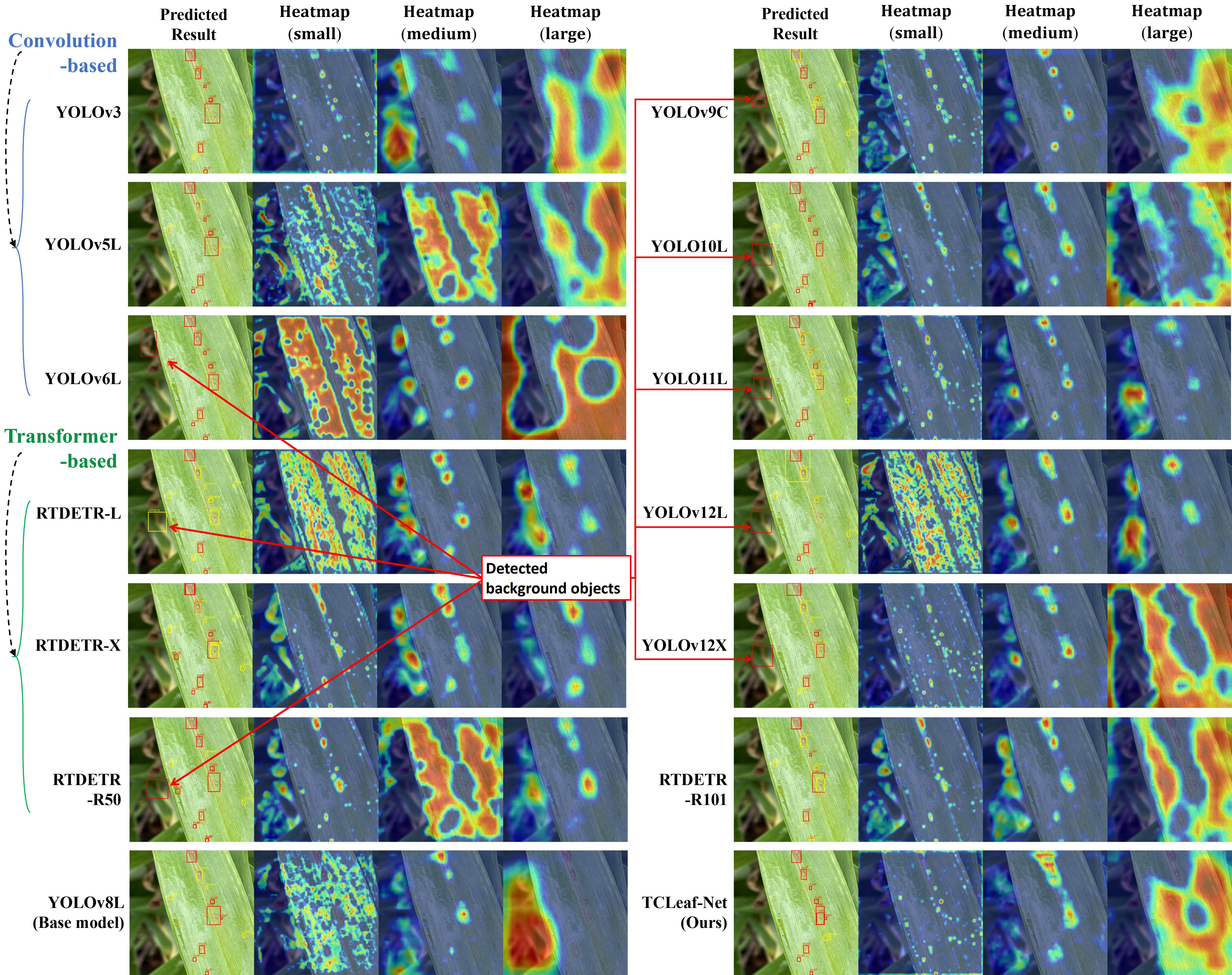}
  \caption{Detection results and decision regions of different models. }
  \label{fig:Fig14}
\end{figure}

Fig. ~\ref{fig:Fig14} contrasts TCLeaf-Net with 13 representative baselines. TCLeaf-Net focuses on lesion pixels, accurately detecting them even when symptoms are very faint, while effectively suppressing background clutter. By contrast, the baselines exhibit diffuse attention that often blankets whole leaves or the surrounding scene, leading to false activations.

This sharp focus stems mainly from the TCM: its three-branch TCL architecture fuses global and local cues, echoing earlier evidence in favor of hybrid attention \citep{Liu2021HAT,Ibtehaz2024ACCViT}. To quantify each branch's contribution, we carried out additional ablations that track how branch removal alters the heatmaps and the final detection scores.

\begin{figure}[htbp]
  \centering
  \includegraphics[width=0.75\linewidth]{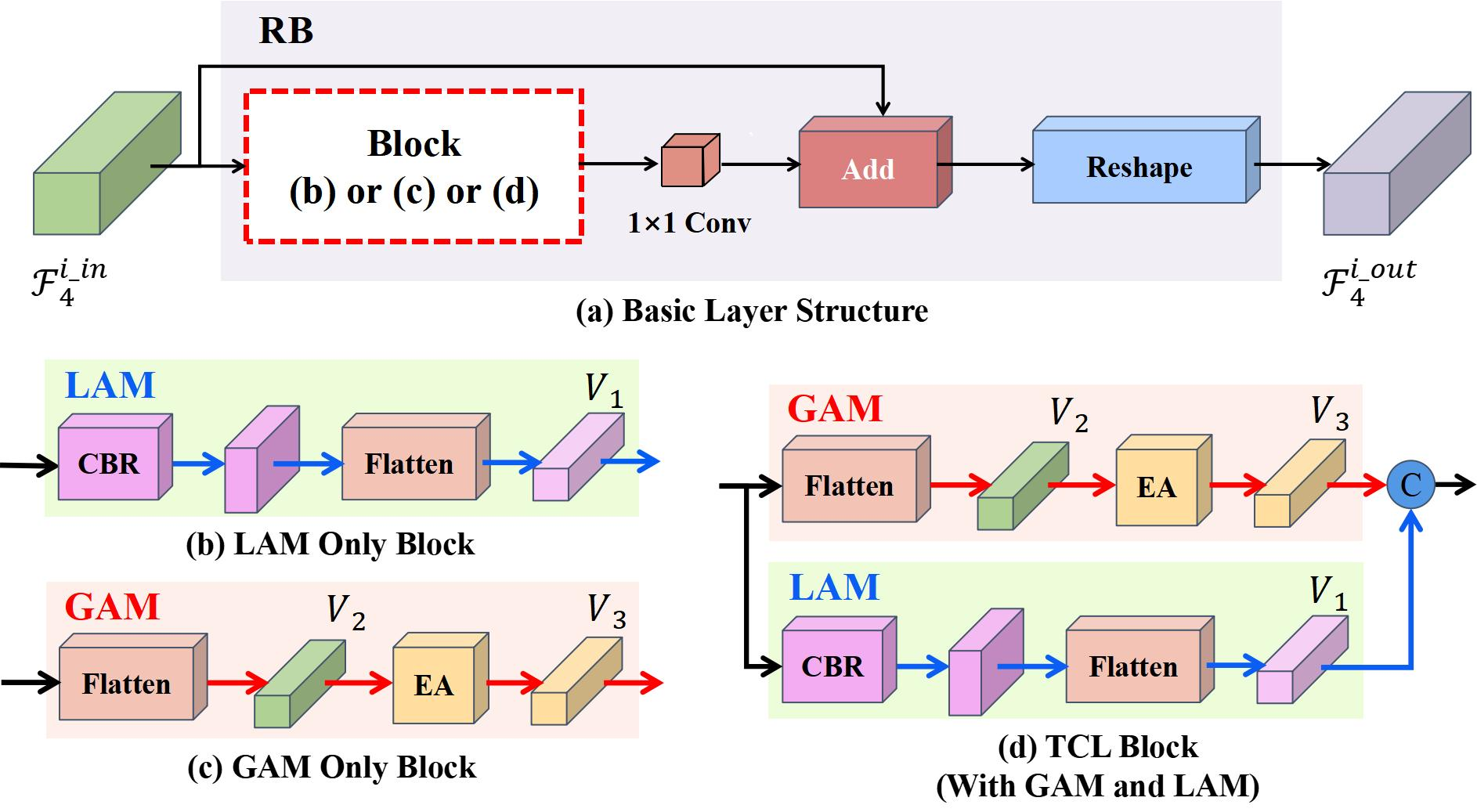}
  \caption{Architectures used in the ablation study of different branches within TCL.}
  \label{fig:Fig15}
\end{figure}

\subsection{Effect of different branches of TCM}

We performed ablation studies on three variants of the TCM: (b) LAM-only Block, which consists solely of the local attention branch (convolution) combined with a residual block (RB); (c) GAM-only Block, which contains only the global attention branch (transformer) with an RB; and (d) TCL Block (this work), which integrates both transformer and convolution branches along with an RB, as illustrated in Fig.~\ref{fig:Fig15}(b)-(d). The basic layer structure is shown in Fig.~\ref{fig:Fig15}(a), with (b), (c), and (d) corresponding to the LAM-only model, GAM-only model, and TCLeaf-Net.

To isolate the benefit of joint global-local processing, we built three variants of the TCM (Fig.~\ref{fig:Fig15}(b)-(d)). LAM-only keeps just the local attention path (convolution) plus a residual block (RB); GAM-only keeps only the global attention path (transformer) and an RB; TCL integrates both paths and the RB.

Table~\ref{tab:t9} clearly demonstrates that TCL (GAM+LAM) achieves the highest recall (69.9\%) among all models, indicating its superior ability to detect true positives in complex scenarios. In addition, TCL also delivers the best mAP@50 (78.2\%) and F1 (77.6\%), while maintaining a high precision (87.2\%).  

In comparison, LAM-only achieves the highest precision (86.8\%) but suffers from a lower recall (63.1\%), and GAM-only provides a moderate improvement in recall (65.3\%) with slightly reduced precision (86.5\%).  

These results highlight that integrating local and global cues in TCL significantly boosts the detection rate without sacrificing overall accuracy.

\begin{table}[ht]\rmfamily
\centering
\caption{Ablation study on each branch of TCL on Daylily-Leaf (in-field) dataset.}
\label{tab:t9}
\resizebox{0.65\linewidth}{!}{%
\begin{tabular}{llccccc}
\toprule
\multicolumn{2}{l}{Branch Scheme} & P (\%) & R (\%) & mAP@50 (\%) & mAP@50:95 (\%) & F1 (\%) \\
\midrule
\multicolumn{2}{l}{LAM-only}   & 86.8  & 63.1  & 74.1  & 53.1  & 73.0 \\
\multicolumn{2}{l}{GAM-only}   & 86.5  & 65.3  & 75.2  & 52.8  & 74.4 \\
\multicolumn{2}{l}{\textbf{GAM+LAM (TCL)}} & \textbf{87.2} & \textbf{69.9} & \textbf{78.2} & \textbf{55.1} & \textbf{77.6} \\
\bottomrule
\end{tabular}
}
\end{table}

\begin{figure}[htbp]
  \centering
  \includegraphics[width=0.85\linewidth]{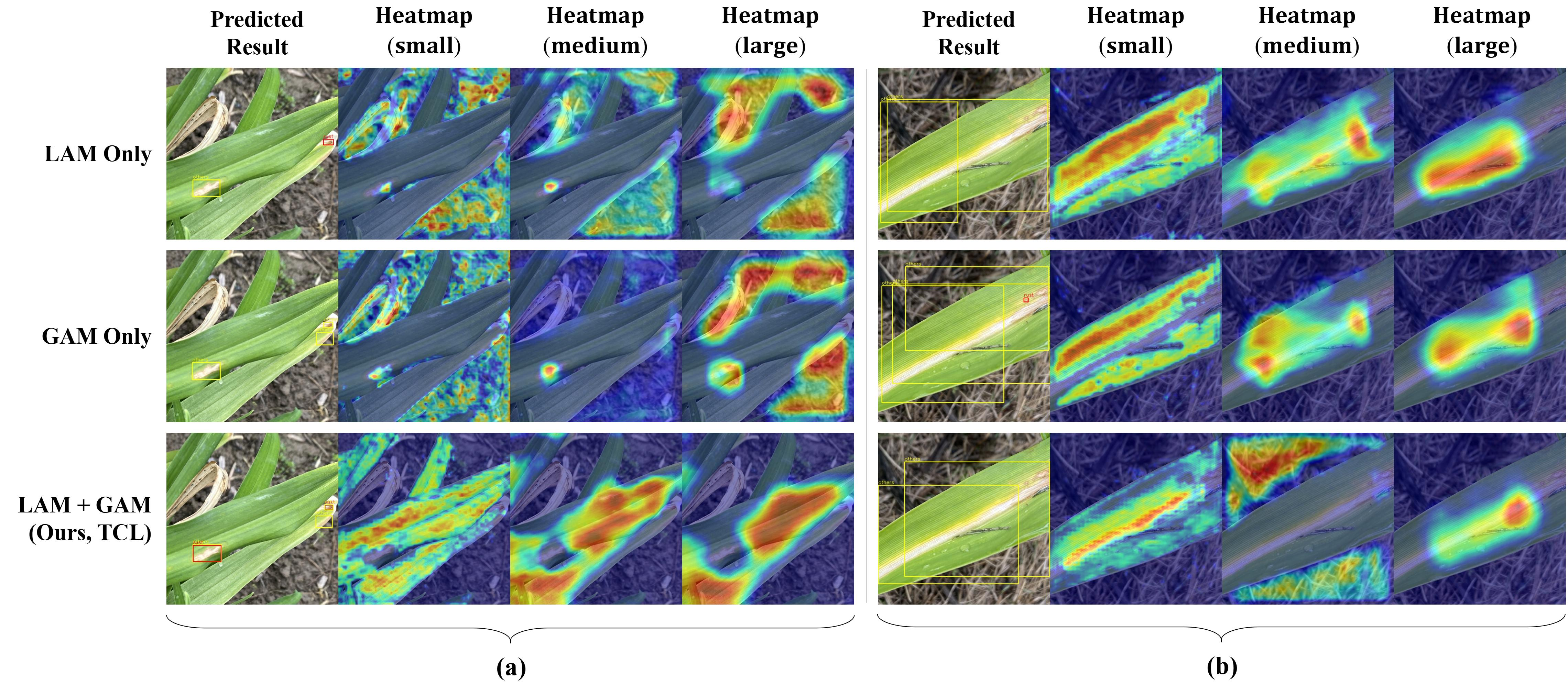}
  \caption{Decision maps generated by GAM, LAM, and TCL for Example 1(a) and Example 2(b).}
  \label{fig:Fig16}
\end{figure}

\begin{figure}[htbp]
  \centering
  \includegraphics[width=0.85\linewidth]{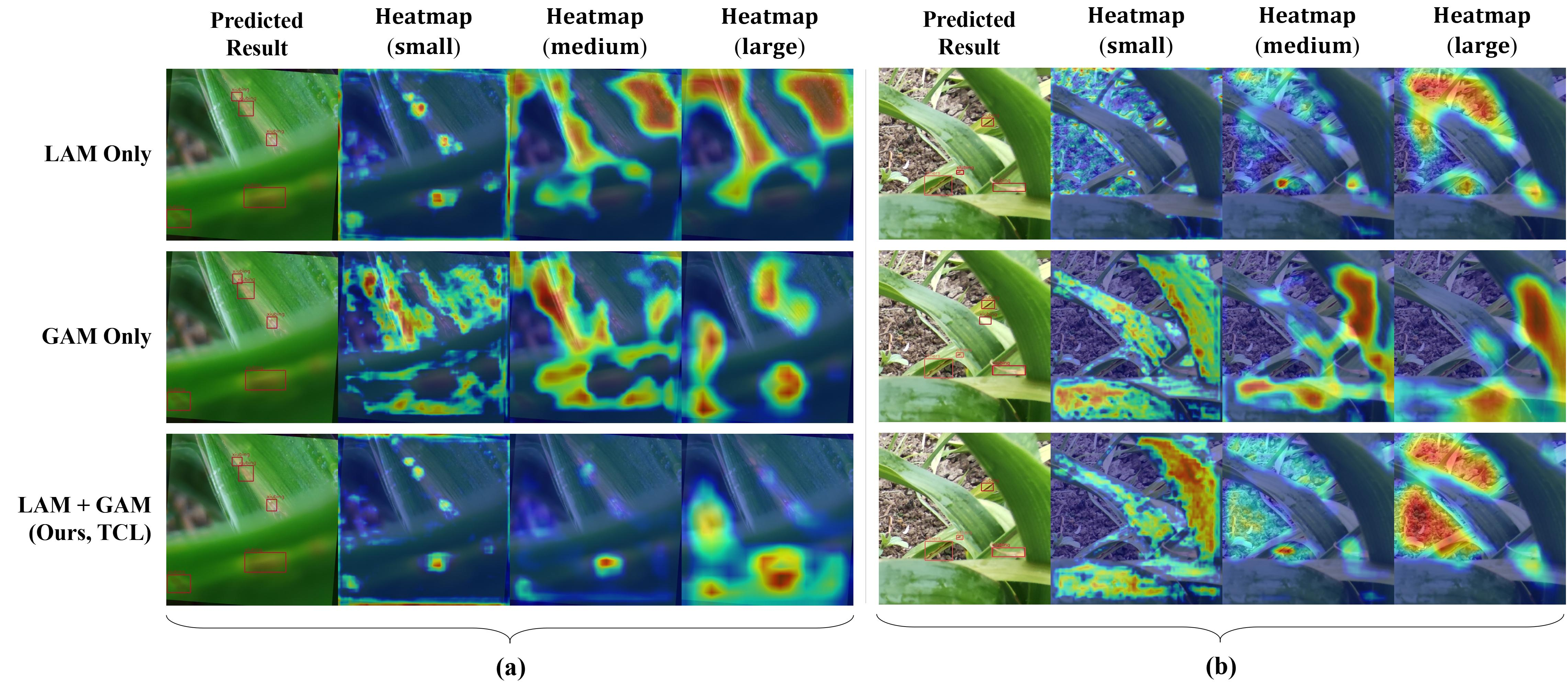}
  \caption{Decision maps generated by GAM, LAM, and TCL for Example 3(a) and Example 4(b).}
  \label{fig:Fig17}
\end{figure}

Grad-CAM heatmaps in Figs.\ref{fig:Fig16} and \ref{fig:Fig17} reinforce the numbers.  
TCL focuses tightly on lesions, GAM-only spreads attention across whole leaves and textured backgrounds, and LAM-only splits focus between lesions and irrelevant pixels.  
For large lesions (Fig.~\ref{fig:Fig16}(a)), TCL covers the symptomatic area most completely, though some lesions still extend beyond a single bounding box, suggesting room for hyper-parameter tuning.

As illustrated in Fig.~\ref{fig:Fig16}(a) and Figs.~\ref{fig:Fig17}(a)-\ref{fig:Fig17}(b), the GAM-only model exhibits \emph{diffuse attention}: its heatmaps spread beyond lesion regions, especially when (i) leaves occupy most of the frame (e.g., Fig.~\ref{fig:Fig17}(a)) or (ii) the background contains rich texture (e.g., Figs.~\ref{fig:Fig16}(a), \ref{fig:Fig17}(b)). In such cases the peak response drifts away from the lesion center, weakening the network's ability to delineate boundaries and increasing the risk of missed detections. This behavior echoes previous observations of global-attention dispersion under clutter \citep{li2024mildetr}.

Combining the LAM branch with the GAM branch restores spatial locality, constraining attention to lesion-relevant regions while complementing the global context modeling provided by GAM. The resulting TCL heatmaps remain lesion-centered and resist background interference, demonstrating stronger robustness under field conditions.

This suggests that controlling both the spatial extent and the central intensity of attention is crucial for reliable disease detection. Although TCLeaf-Net significantly improves attention focus and suppresses responses to background clutter, its attention intensity on lesion centers does not always surpass that of LAM-only in certain scenes, such as Fig.~\ref{fig:Fig17}(a). Future work may focus on adaptive attention gating strategies to further enhance lesion-centered focus without compromising robustness.

In parallel with improving attention precision, computational efficiency is another key consideration in real-world deployments. Within TCM, the GAM branch accounts for most of the computational overhead, as global attention is commonly implemented using MHSA. To reduce resource consumption, we employ efficient attention (EA), which maintains global modeling capability with lower cost \citep{qin2023factorization,ma2024semantic}. Given EA's central role in the GAM branch, we further investigate its standalone effectiveness below.

\subsection{Effectiveness of efficient attention (EA)}
To evaluate the role of EA in TCLeaf-Net, we replaced it with two commonly used global attention mechanisms: MHSA and cross-attention (CA). Table~\ref{tab:t10} summarizes the results.

EA achieves the highest mAP@50 and F1 among the three, while matching MHSA in recall and slightly surpassing it in precision. Compared to CA, which yields the best precision (87.3\%), EA maintains nearly equivalent precision (87.2\%) but gains a 2.6\% recall advantage, leading to the best F1 (77.6\%). This suggests that EA is more balanced, capturing additional true positives without significantly increasing false positives.

In terms of efficiency, EA requires only 9.4~GB of GPU memory-less than half of what MHSA and CA consume-and reduces total training time by over 37\%. MHSA and CA also exhibit out-of-memory (OOM) issues on 24~GB cards in training, limiting their practicality for mid-range GPUs.

These results highlight EA's favorable trade-off between accuracy and efficiency. It supports effective global feature extraction under limited computational budgets, making TCLeaf-Net deployable on consumer-grade GPUs such as the RTX~3060 (12~GB), without compromising performance in complex agricultural scenarios.

\begin{table}[ht]\rmfamily
\centering
\caption{Comparison of global attention mechanisms on the Daylily-Leaf (in-field) dataset.}
\label{tab:t10}
\resizebox{0.7\linewidth}{!}{%
\begin{tabular}{llllllll}
\toprule
Type & GPU (GB)$\downarrow$ & Time (h)$\downarrow$ & 
P (\%)$\uparrow$ & R (\%)$\uparrow$ & 
mAP@50 (\%)$\uparrow$ & mAP@50:95 (\%)$\uparrow$ & F1 (\%)$\uparrow$ \\
\midrule
MHSA* & $\sim$25.0 & 1.04 & 85.4 & \textbf{69.9} & 77.2 & 55.8 & 76.9 \\
CA*   & $\sim$26.1 & 1.06 & \textbf{87.3} & 67.3 & 78.0 & \textbf{56.8} & 76.0 \\
\textbf{EA}   & \textbf{9.4} & \textbf{0.66} & 87.2 & \textbf{69.9} & \textbf{78.2} & 55.1 & \textbf{77.6} \\
\bottomrule
\end{tabular}
}

\vspace{1pt}
\par
\raggedright
\footnotesize
\textbf{Note.}
\textbf{Type} = Attention mechanism; \quad
\textbf{GPU} = Memory usage during training (GB); \quad
\textbf{Time} = Total training time (hours); \\
\textbf{*} Due to out-of-memory (OOM) on the RTX 4090D (24GB), training was conducted on a 32GB GPU.
\end{table}

Beyond controlled evaluations, field conditions present unpredictable challenges. We therefore examine TCLeaf-Net's robustness under various synthetic distortions.

\begin{figure}[htbp]
  \centering
  \includegraphics[width=0.85\linewidth]{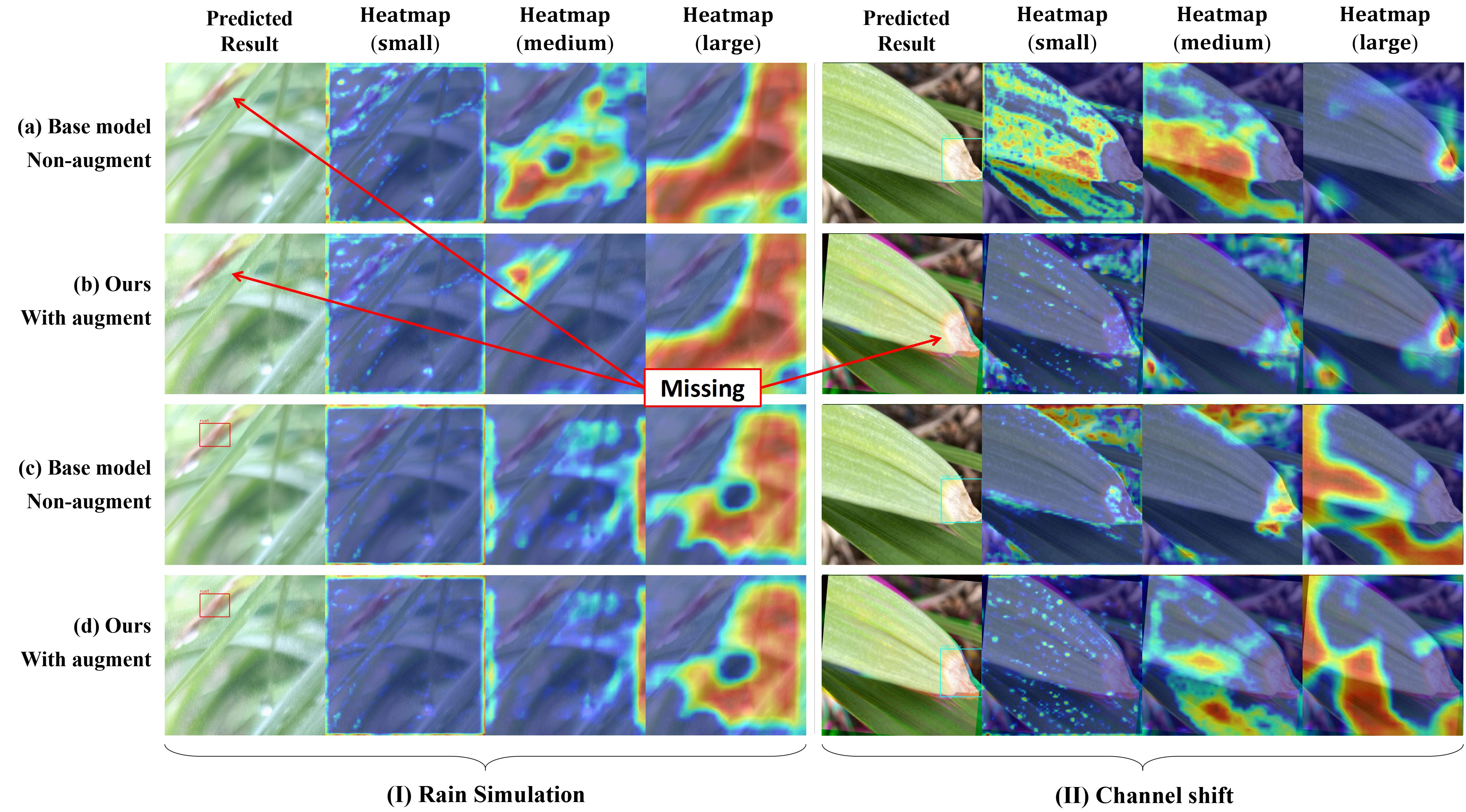}
  \caption{Comparison of detection results and decision maps under (a) simulated rainy conditions and (b) simulated jitter, channel shift conditions.}
  \label{fig:Fig18}
\end{figure}

\begin{figure}[htbp]
  \centering
  \includegraphics[width=0.65\linewidth]{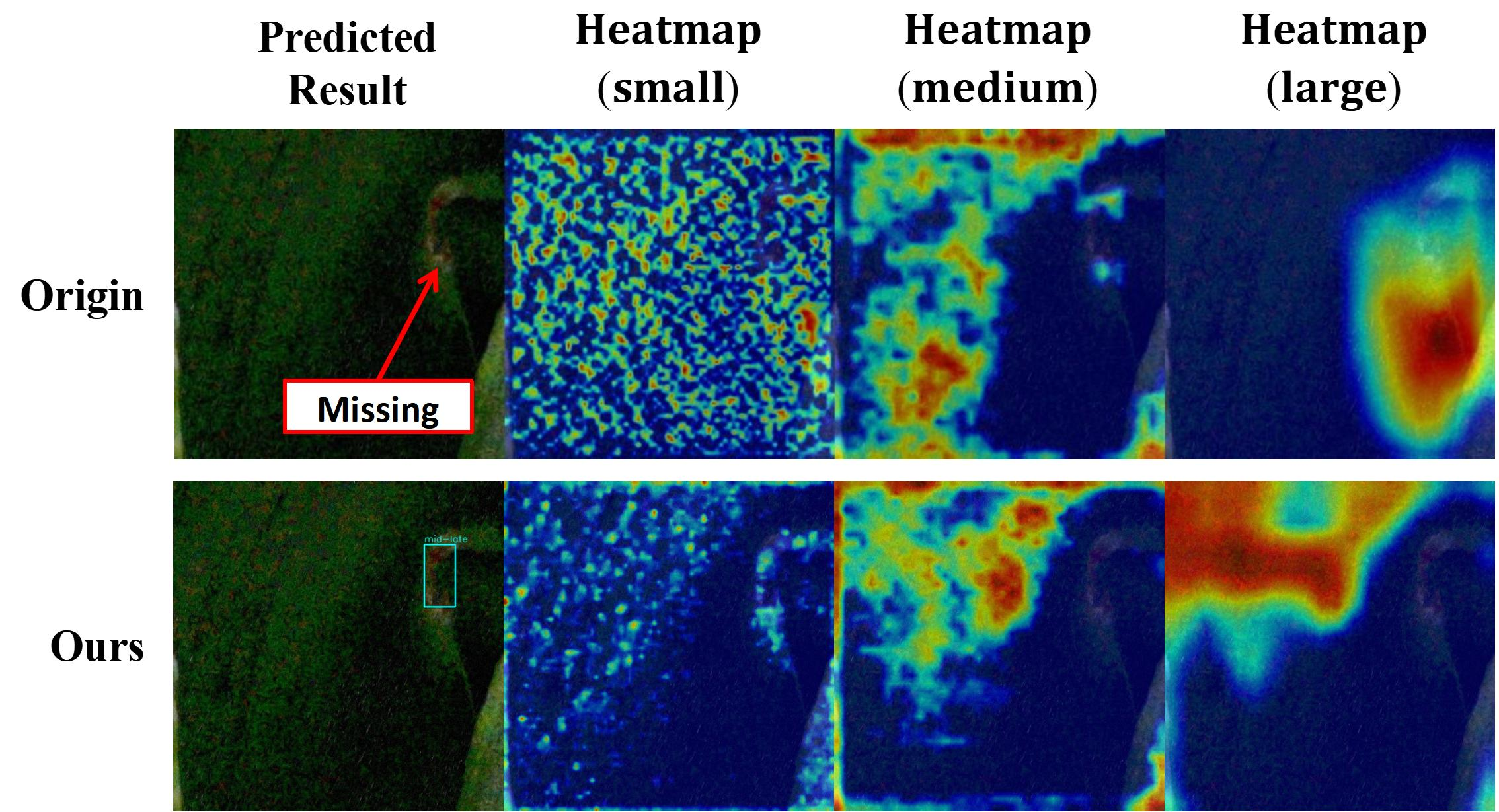}
  \caption{Comparison of detection results and decision maps in low-light and noisy complex environments.}
  \label{fig:Fig19}
\end{figure}

\subsection{Robustness of TCLeaf-Net against environmental variations}
To test the TCLeaf-Net's real-world resilience, we generated three types of synthetic perturbations: rain streaks that emulate adverse weather, a compound rotation with channel shifts that reproduces camera shake, and low-light sensor noise typical of night-time photography. Figs. \ref{fig:Fig18}-\ref{fig:Fig19} show that TCLeaf-Net remains accurate under each condition, whereas the base model either misses lesions or misclassifies background regions.

Under rainy condition (Fig.~\ref{fig:Fig18}(a)), TCLeaf-Net still localizes the target, whereas the base model misses it entirely. After rotation and color shift (Fig.~\ref{fig:Fig18}(b)) the baseline fails again, but TCLeaf-Net preserves tight boxes. In dim, noisy scenes (Fig.~\ref{fig:Fig19}) TCLeaf-Net identifies lesions correctly, while the baseline misclassifies background patches. These examples underline the model's robustness to typical field-level distortions.

\subsection{Limitations}

While TCLeaf-Net demonstrates strong performance across multiple datasets and perturbation scenarios, several limitations remain that warrant further investigation.

First, although the transformer-convolution module (TCM) fuses global and local representations via $1 \times 1$ convolutions, the fusion weights are spatially uniform and shared across positions. This may limit the model's ability to dynamically adjust feature contributions across different lesion sizes or background conditions. In some cases, transformer-dominant responses may produce overly diffuse attention maps, while convolution-dominant responses may miss long-range dependencies.

Second, the current design of the multi-scale detection heads is strictly hierarchical: features from different levels are assigned to distinct heads without cross-level refinement. This structure can lead to fragmented predictions when lesions of different scales overlap or when small lesions appear near large ones. While DFPN improves alignment, its scale routing remains fixed. Introducing inter-head communication or scale-aware fusion may help address this issue.

Third, the Daylily-Leaf dataset, although annotated at lesion level and collected under diverse conditions, includes only one plant species and a limited number of symptom categories. This restricts the scope of generalization, especially for models intended for multi-crop deployment. Some complex or mixed symptoms are also coarsely labeled due to annotation constraints.

Future work will address these issues by incorporating spatially dynamic fusion in the backbone, refining cross-scale feature interactions, and expanding the dataset to include more crop species, a wider variety of disease symptoms, and different stages of plant growth with finer annotation granularity.

\section{Conclusion}

This study presents TCLeaf-Net, a transformer-convolution hybrid architecture designed for robust lesion-level plant disease detection under complex in-field conditions. To support this objective, we construct Daylily-Leaf (ideal and in-field), a pair of lesion-level annotated datasets comprising fine-grained bounding boxes collected under both laboratory and real-world environments. Despite their compact scale, the datasets capture representative lesion patterns and background variability, providing a practical benchmark for instance-level plant disease detection in the wild.

Through extensive experiments, we observe that pure transformer-based detectors are prone to diffuse attention in cluttered backgrounds, likely due to their weak inductive biases and the lack of spatial priors. By integrating a convolutional branch, TCLeaf-Net enhances spatial focus on leaf and lesion regions, effectively suppressing false positives arising from complex visual scenes.

In addition to its hybrid backbone, TCLeaf-Net incorporates several architectural innovations to enhance feature representation and alignment. The SSOPE and RSFRS modules help preserve fine-grained spatial cues during early downsampling. The DFPN, equipped with multi-scale receptive fields and adaptive alignment, improves localization and semantic integration across scales.

On Daylily-Leaf, TCLeaf-Net surpasses all 13 CNN- and transformer-based baselines, reaching \(\mathrm{mAP}@50 = 89.5\% / 78.2\%\) (Ideal / In-Field) with \(46.1\,\mathrm{M}\) parameters and \(157.9\,\mathrm{GFLOPs}\). On external leaf-level detection benchmarks (PlantDoc, Tomato-Leaf, Rice-Leaf), it attains \(\mathrm{mAP}@50 = 65.7\%, 94.6\%, 57.3\%\), respectively, demonstrating strong generalization and a practical computational footprint for deployment in diverse agricultural scenarios.

This study has two contributions:
\begin{itemize}
    \item A small-scale yet carefully constructed lesion-level dataset, with rich annotations across varied lesion types and backgrounds, enabling fine-grained disease detection in real-world conditions;
    \item A hybrid detection framework that effectively addresses core limitations of existing CNN and transformer architectures under field conditions through improved feature preservation and robust background suppression.
\end{itemize}

These contributions lay the groundwork for further advancements in crop disease detection. Future work will focus on scaling the dataset to include more crop species and symptom types, optimizing model efficiency for mobile deployment, and exploring semi-supervised learning strategies to reduce annotation costs while maintaining lesion-level accuracy.






\section*{Declaration of competing interest}
The authors declare that they have no known competing financial interests or personal relationships
that could have appeared to influence the work reported in this paper.

\section*{Data availability}
Data will be made available on reasonable request.

\printcredits

\bibliographystyle{cas-model2-names}

\bibliography{cas-refs_origin}






\end{document}